\documentclass[10pt,twocolumn,letterpaper]{article}

\usepackage{times}
\usepackage{epsfig}
\usepackage{graphicx}
\usepackage{amsmath}
\usepackage{amssymb}
\usepackage{amsthm}
\usepackage{booktabs}
\usepackage{multirow}
\usepackage{graphicx}
\usepackage{algorithm}
\usepackage{algorithmic}
\usepackage{cite}
\usepackage{wrapfig}
\usepackage{caption}
\usepackage[normalem]{ulem}

\usepackage{authblk}
\usepackage{appendix}

\newcommand{\coolname}{$\texttt{InCTRL}$\xspace}

\usepackage[pagenumbers]{cvpr} 

\definecolor{cvprblue}{rgb}{0.21,0.49,0.74}

\usepackage[pagebackref,breaklinks,colorlinks,citecolor=cvprblue]{hyperref}

\def\confName{CVPR}
\def\confYear{2024}

\title{Toward Generalist Anomaly Detection via In-context Residual Learning with Few-shot
Sample Prompts}

\author{Jiawen Zhu}
\author{Guansong Pang \thanks{Corresponding author: G. Pang (\tt\small gspang@smu.edu.sg)}}
\affil{School of Computing and Information Systems, Singapore Management University}

\begin{document}
\maketitle
\begin{abstract}

This paper explores the problem of Generalist Anomaly Detection (GAD), aiming to train one single detection model that can generalize to detect anomalies in diverse datasets from different application domains without any further training on the target data.
Some recent studies have shown that large pre-trained Visual-Language Models (VLMs) like CLIP have strong generalization capabilities on detecting industrial defects from various datasets,
but their methods rely heavily on handcrafted text prompts about defects, making them difficult to generalize to anomalies in other applications, \eg, medical image anomalies or semantic anomalies in natural images. 
In this work, we propose to train a GAD model with few-shot normal images as sample prompts for AD on diverse datasets on the fly. To this end, we introduce a novel approach that learns an \underline{in}-\underline{c}on\underline{t}ext \underline{r}esidual \underline{l}earning model for GAD, termed \coolname.
It is trained on an auxiliary dataset to discriminate anomalies from normal samples based on a holistic evaluation of the residuals between query images and few-shot normal sample prompts. Regardless of the datasets, per definition of anomaly, larger residuals are expected for anomalies than normal samples, thereby enabling \coolname to generalize across different domains without further training.
\begin{figure}[t!]
    \centering
    \includegraphics[width=0.45\textwidth]{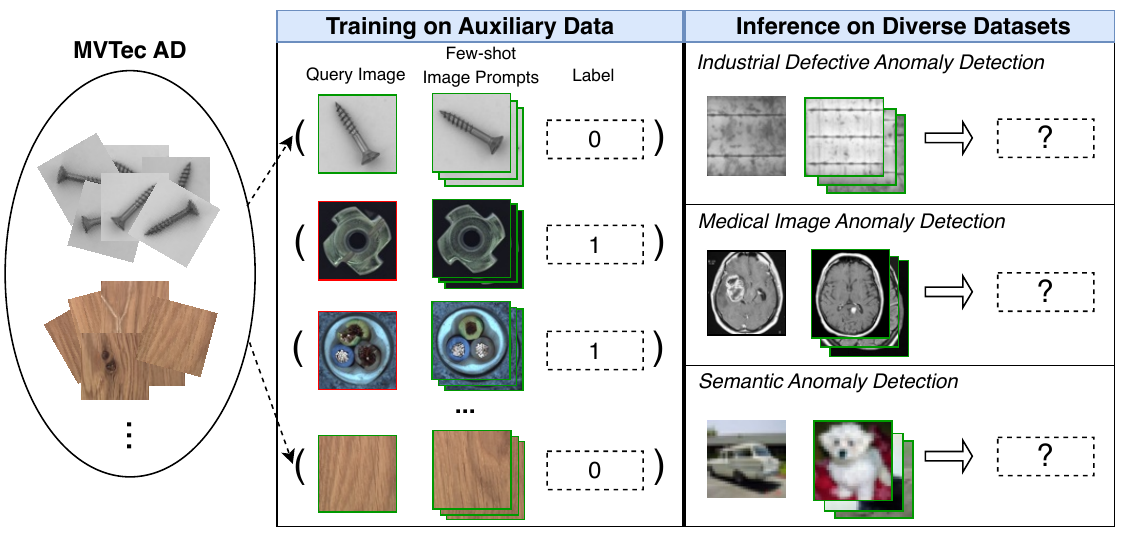}
    \includegraphics[width=0.45\textwidth]{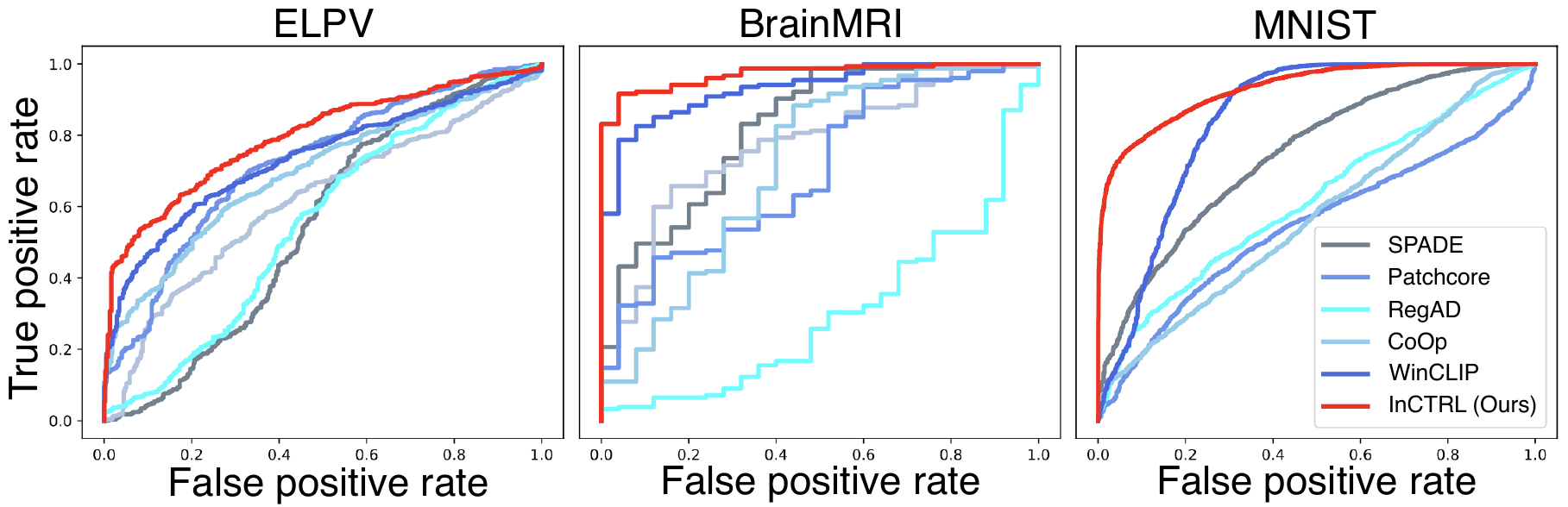}
    \caption{\textbf{Top}: An illustration of \coolname: a one-for-all model using few-shot normal images as sample prompts. \textbf{Bottom}: AUROC curves of \coolname and competing few-shot methods on three different application datasets without any training on the target data.} 
    \label{fig:intro}
    \vspace{-0.5cm}
\end{figure}
Comprehensive experiments on nine AD datasets are performed to establish a GAD benchmark that encapsulate the detection of industrial defect anomalies, medical anomalies, and semantic anomalies in both one-vs-all and multi-class setting, on which \coolname is the best performer and significantly outperforms state-of-the-art competing methods. Code is available at \renewcommand\UrlFont{\color{blue}}\url{https://github.com/mala-lab/InCTRL}.

\end{abstract}
\vspace{-0.3cm}
\section{Introduction}

Anomaly Detection (AD) is a crucial computer vision task that aims to detect samples that substantially deviate from the majority of samples in a dataset, due to its broad real-life applications such as industrial inspection, medical imaging analysis, and scientific discovery, etc. \cite{pang2021deep,cao2024survey}. 
Current AD paradigms are focused on individually building one model on the training data, \eg, a set of anomaly-free samples, of each target dataset, such as data reconstruction approach~\cite{akcay2019ganomaly, schlegl2019f, zavrtanik2021reconstruction, yan2021learning, zaheer2020old, zavrtanik2021draem, park2020learning, hou2021divide, xiang2023squid, liu2023diversity, yao2023one, yao2023focus}, one-class classification~\cite{tax2004support, yi2020patch, bergman2020classification,chen2022deep,ruff2020deep}, and knowledge distillation approach~\cite{deng2022anomaly, bergmann2020uninformed, salehi2021multiresolution, wang2021student, Cao_2023_ICCV, tien2023revisiting, zhang2023destseg}.

Although these approaches have shown remarkable detection performance on various AD benchmarks, they require the availability of large training data and the skillful training of the detection model per dataset. Thus, they become infeasible in application scenarios where training on the target dataset is not allowed due to either data privacy issues, \eg, arising from using those data in training the models due to machine unlearning \cite{xu2023machine}, or unavailability of large-scale training data in the deployment of new applications. To tackle these challenges, this paper explores the problem of learning \textbf{Generalist Anomaly Detection} (GAD) models, \textit{aiming to train one single detection model that can generalize to detect anomalies in diverse datasets from different application domains without any training on the target data}.

Being pre-trained on web-scale image-text data, large Visual-Language Models (VLMs) like CLIP \cite{radford2021learning} have exhibited superior generalization capabilities in recent years, achieving accurate visual recognition across different datasets without any fine-tuning or adaptation on the target data. More importantly, some very recent studies (\eg, WinCLIP \cite{jeong2023winclip}) show that these VLMs can also be utilized to achieve remarkable generalization on different defect detection datasets. Nevertheless, a significant limitation of these models is their dependency on a large set of manually crafted prompts specific to defects. This reliance restricts their applicability, making it challenging to extend their use to detecting anomalies in other data domains, \eg, medical image anomalies \cite{salehi2021multiresolution,ding2022catching,tian2021constrained,Cai_2023,tian2023self} or semantic anomalies in one-vs-all or multi-class settings \cite{Cao_2023_ICCV,ruff2020deep}.

To address this problem, we propose to train a GAD model that aims to utilize few-shot normal images from any target dataset as sample prompts for supporting GAD on the fly, as illustrated in Figure \ref{fig:intro}(Top). The few-shot setting is motivated by the fact that it is often easy to obtain few-shot normal images in real-world applications. Furthermore, these few-shot samples are not used for model training/tuning; they are just used as sample prompts for enabling the anomaly scoring of test images during inference. This formulation is fundamentally different from current few-shot AD methods~\cite{sheynin2021hierarchical, huang2022registration, wu2021learning, belton2023fewsome, schwartz2022maeday, wang2022few, xie2023pushing} that use these target samples and their extensive augmented versions to train the detection model, which can lead to an overfitting of the target dataset and fail to generalize to other datasets, as shown in Figure \ref{fig:intro}(Bottom). 

We then introduce an GAD approach, the first of its kind, that learns an \underline{in}-\underline{c}on\underline{t}ext \underline{r}esidual \underline{l}earning model based on CLIP, termed \coolname. It trains an GAD model to discriminate anomalies from normal samples by learning to identify the residuals/discrepancies between query images and a set of few-shot normal images from auxiliary data. The few-shot normal images, namely \textit{in-context sample prompts}, serve as prototypes of normal patterns. When comparing with the features of these normal patterns, per definition of anomaly, a larger residual is typically expected for anomalies than normal samples in datasets of different domains, so the learned in-context residual model can generalize to detect diverse types of anomalies across the domains.

To capture the residuals better, \coolname models the in-context residuals at both the image and patch levels, gaining an in-depth in-context understanding of what constitutes an anomaly. Further, our in-context residual learning can also enable a seamless incorporation of normal/abnormal text prompt-guided prior knowledge into the detection model, providing an additional strength for the detection from the text-image-aligned semantic space.

Accordingly, we make the following main contributions.
\begin{itemize}
\item We introduce a GAD task to evaluate the generalization capability of AD methods in identifying anomalies across various scenarios without needing to training/tuning on the target datasets. To the best of our knowledge, this is the first study dedicated to a generalist approach to anomaly detection, encompassing industrial defects, medical anomalies, and semantic anomalies.
\item We then propose an in-context residual learning framework for GAD, called \coolname. It is designed to distinguish anomalies from normal samples by detecting residuals between test images and in-context few-shot normal sample prompts from any target dataset on the fly. \coolname is optimized on auxiliary data to achieve the one-model-for-all goal, \ie, one model for AD on diverse datasets without any training on target data.
\item Comprehensive experiments on nine diverse AD datasets are performed to establish a GAD benchmark that encapsulates three types of popular AD tasks, including industrial defect anomaly detection, medical image anomaly detection, and semantic anomaly detection under both one-vs-all and multi-class settings. Our results show that \coolname significantly outperforms state-of-the-art competing methods.
\end{itemize}

\section{Related Work}
\subsection{Anomaly Detection}
\vspace{0.1cm}
\noindent\textbf{Anomaly Detection.}
Existing AD approaches typically rely on unsupervised learning due to the scarcity of anomaly data. Numerous methods have been introduced. One-class classification methods~\cite{tax2004support, yi2020patch, bergman2020classification,chen2022deep,ruff2020deep} focus on compactly describing normal data with support vectors. Reconstruction-based methods~\cite{akcay2019ganomaly, schlegl2019f, zavrtanik2021reconstruction, yan2021learning, zaheer2020old, zavrtanik2021draem, park2020learning, hou2021divide, xiang2023squid, liu2023diversity, yao2023one, yao2023focus} train models to reconstruct normal images, where anomalies are identified by higher reconstruction errors. Distance-based methods~\cite{defard2021padim, cohen2020sub, roth2022towards} determine anomalies based on the distance between test image embeddings and normal reference embeddings from stored training data. Knowledge distillation methods~\cite{deng2022anomaly, bergmann2020uninformed, salehi2021multiresolution, wang2021student, Cao_2023_ICCV, tien2023revisiting, zhang2023destseg} focus on distilling normal patterns from pre-trained models and detect anomalies based on the difference between distilled and original features. The above approaches are designed to fit on target dataset for AD, \ie, one model for one dataset. We aim for a one-model-for-all setting. A relevant research line is to tackle the AD problem under domain or distribution shift~\cite{aich2023cross,yao2023one,lu2020few,Cao_2023_ICCV,zhu2023anomaly,ding2022catching}, but they generally assume a large domain relevance on the source and target data. Additionally, there have been a number of concurrent studies leveraging VLMs for AD \cite{wu2023open,wu2023vadclip,zhou2024anomalyclip}, but they address a different setting from ours, \eg, weakly-supervised AD \cite{wu2023open,wu2023vadclip} or zero-shot AD \cite{zhou2024anomalyclip}.

\vspace{0.1cm}
\noindent\textbf{Few-shot Anomaly Detection (FSAD).}
FSAD is designed to identify anomalies using only a limited number of normal samples from target datasets. Traditional FSAD research focuses on modeling the normal distribution of these few normal samples to detect anomalies~\cite{sheynin2021hierarchical, huang2022registration, wu2021learning, belton2023fewsome, schwartz2022maeday, wang2022few, xie2023pushing,liao2024coft}. However, these methods often cannot generalize to new domains, as they generally require re-training or fine-tuning with normal data from the target datasets.

Distance-based approaches such as SPADE~\cite{cohen2020sub}, PaDiM~\cite{defard2021padim} and PatchCore~\cite{roth2022towards} present a solution to address this problem by making full use of available pre-trained representations of the few-shot samples to calculate distance-based anomaly scores without training. Recently, RegAD~\cite{huang2022registration} is designed as a model that operates without the need for training or fine-tuning on new data for the FSAD task, but it requires domain relevance between training and test data to work well. WinCLIP~\cite{jeong2023winclip} pioneers the application of large Visual-Language Models (VLM) on zero-shot and few-shot anomaly detection tasks by processing images through multi-scale window movements and text prompting to CLIP. Without adapting CLIP to the AD task, WinCLIP gains impressive zero-shot detection performance on defect datasets using its handcrafted text prompts, but it fails to work well when the text prompts cannot capture the required anomaly semantics, making it difficult to generalize well to diverse anomaly detection tasks. 

\subsection{In-Context Learning}

In-context learning is an innovative approach that helps enhance the performance of Large Language Models (LLMs) in Natural Language Processing (NLP)~\cite{alayrac2022flamingo, brown2020language, hao2022language}, which leverages minimal in-context prompts to adapt LLMs to novel tasks effectively. 

Recently, several studies \cite{chen2021pix2seq, chen2022unified, kolesnikov2022uvim, lu2022unified, wang2022ofa} attempt to apply in-context learning to vision tasks by converting vision problems to NLP ones using the language or specially-designed discrete tokens as the task prompts. On the other hand, Amir \textit{et al.} \cite{bar2022visual} introduce a novel approach for in-context visual prompting by treating a spectrum of vision tasks as grid in-painting problems. Similarly, Painter \cite{wang2023images, wang2023seggpt} then proposes to perform masked image in-painting. However, these methods focus more on task-level generalization, so they are not applicable to the AD task which focuses more on the instance-level discrepancy.

Our work redesign in-context learning for GAD. We redefine image prompts as dataset-specific normal patterns, rather than as an instruction for particular tasks. By capturing the in-context residual between the query image and few-shot normal prompts, our model can gain a cohesive understanding of diverse anomalies, enabling remarkable generalized detection performance for GAD.

\begin{figure*}[ht]
    \centering
    \includegraphics[width=0.92\textwidth]{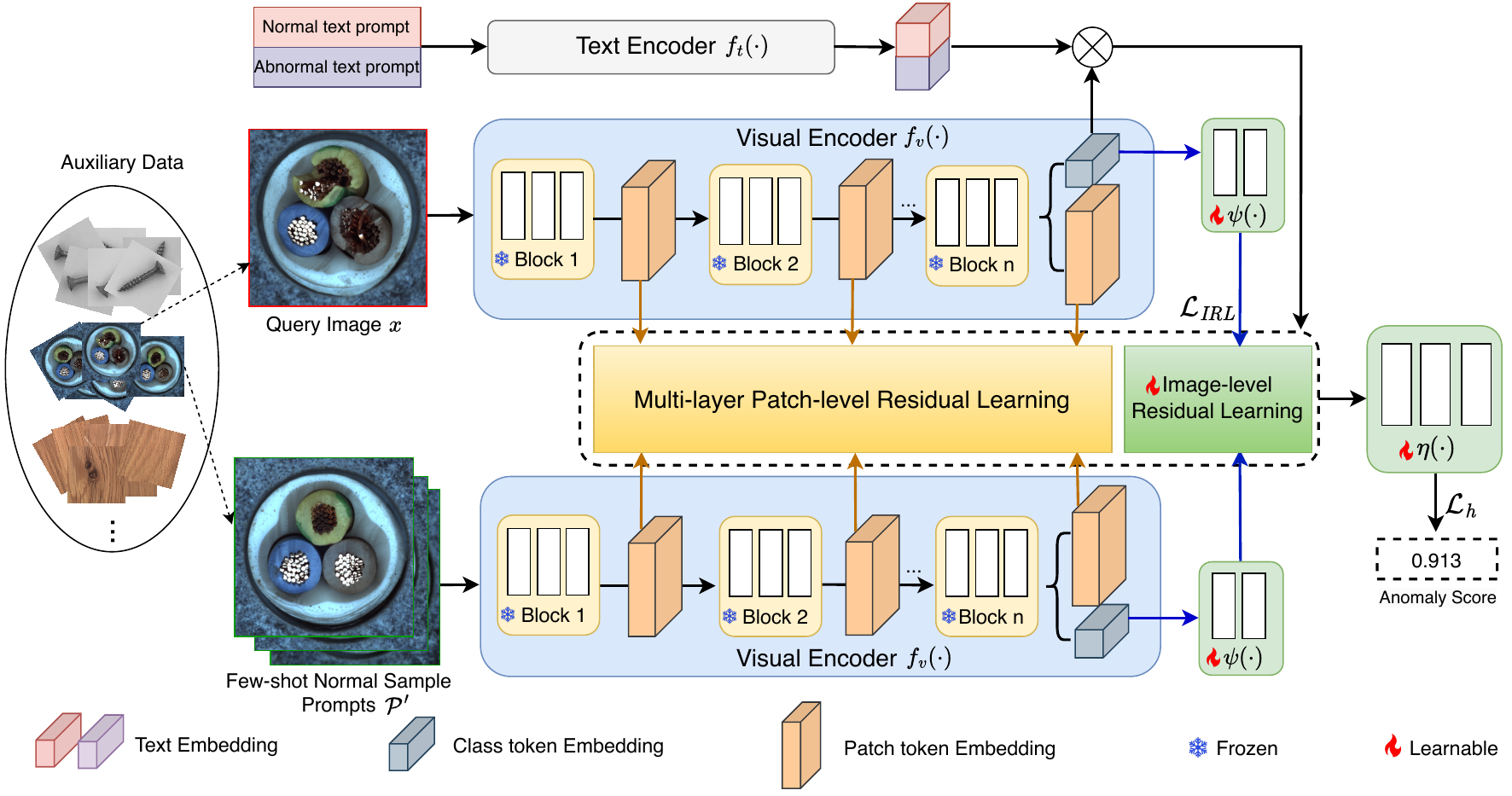}
    \caption{Overview of the training of \coolname. It firstly simulates in-context learning scenarios using a query image and a few-shot normal sample prompts randomly drawn from the auxiliary training data. Then it performs multi-layer patch-level and image-level residual learning to capture both local and global residuals between the query image and the normal prompts. Lastly, those residual information, combined with text prompts-guided prior knowledge from the text encoder, is utilized for a holistic anomaly score learning.}
    \label{fig:overall}
    \vspace{-0.3cm}
\end{figure*}

\section{\coolname: In-Context Residual Learning}

\subsection{Problem Statement}
The objective of GAD is to train a single AD model that works well for detecting anomalies on test datasets from diverse application domains without any training on the target data. Thus, the training set is assumed to be drawn from different distributions from the test sets.
Formally, let $\mathcal{D}_{train} = \{X_{train}, Y_{train}\}$ be an \textit{auxiliary} training dataset with normal and anomaly class labels, where $X_{train} = \{x_i\}_{i=1}^{N}$ consists of $N$ normal and anomalous images and $Y_{train} = \{y_i\}_{i=1}^{N}$, with $y_i=0$ indicates normal and $y_i=1$ signifies abnormal.
A collection of test sets, $\mathcal{T}=\{\mathcal{D}_{test}^1,\mathcal{D}_{test}^2,\cdots,\mathcal{D}_{test}^M\}$ with $\mathcal{D}_{test}^j = \{X_{test}^j, Y_{test}^j\}$, from $M$ different application domains with various types of anomalies is given. The test sets are drawn from a distribution different from that of $\mathcal{D}_{train}$. Then the goal is to train a generalist anomaly scoring function$: \mathcal{D}_{train} \rightarrow \mathbb{R}$ so that it assigns larger anomaly scores to the anomalous samples than to the normal ones from any test dataset in $\mathcal{T}$.
In the context of GAD with few-shot normal samples, a small set of a few normal images randomly drawn from the target domain, $\mathcal{P}=\{p_1, p_2, \cdots,p_K\}$ where $K$ is typically a small number, \eg, $K \ll N$, is available during inference, but $\mathcal{P}$ is not available in any way during the training of the generalist detection model

\subsection{Overview of Our Approach \textbf{\coolname}}
Our approach \coolname is designed to effectively model the in-context residual between a query image and a set of few-shot normal images as sample prompts, utilizing the generalization capabilities of CLIP to detect unusual residuals for anomalies from different application domains.
CLIP is a VLM consisting of a text encoder $f_t(\cdot)$ and a visual encoder $f_v(\cdot)$, with the image and text representations from these encoders well aligned by pre-training on web-scale text-image data. \coolname is optimized using auxiliary data $\mathcal{D}_{train}$ via an in-context residual learning in the image encoder, with the learning augmented by text prompt-guided prior knowledge from the text encoder. 

To be more specific, as illustrated in Fig.~\ref{fig:overall}, we first simulate an in-context learning example that contains one query image $x$ and a set of few-shot normal sample prompts $\mathcal{P}^\prime$, both of which are randomly sampled from the auxiliary data $\mathcal{D}_{train}$. Through the visual encoder, we then perform multi-layer patch-level and image-level residual learning to respectively capture local and global discrepancies between the query and few-shot normal sample prompts (Secs. \ref{subsec:patch} and \ref{subsec:image}). 
Further, our model allows a seamless incorporation of normal and abnormal text prompts-guided prior knowledge from the text encoder based on the similarity between these textual prompt embeddings and the query images (Sec. \ref{subsec:text}). The training of \coolname is to optimize a few projection/adaptation layers attached to the visual encoder to learn a larger anomaly score for anomaly samples than normal samples in $\mathcal{D}_{train}$, with the original parameters in both encoders frozen; during inference, a test image, together with the few-shot normal image prompts from the target dataset and the text prompts, is put forward through our adapted CLIP-based GAD network, whose output is the anomaly score for the test image (Sec. \ref{subsec:train_inference}). Below we present these modules in detail.

\subsection{Multi-Layer Patch-Level Residual Learning}\label{subsec:patch}
To effectively capture fine-grained in-context residuals between the query image and the normal image prompts, we introduce a multi-layer patch-level residual learning component in \coolname. Typically, the CLIP visual encoder comprises a series of block layers. From the bottom to the top of layers, the visual encoder gradually learns the visual patterns at different levels of abstraction~\cite{radford2021learning}. Thus, this component is designed to model patch-level in-context residuals from the patch token embeddings obtained from the multiple levels of the blocks within the visual encoder.
 
To be specific, assuming the visual encoder consists of $n$ blocks, for a given set of few-shot normal sample prompts $\mathcal{P}^\prime$ and a training query image $x$, we extract a series of patch token embedding maps $\{T_x^l\}_{l=1}^n$ and $\{T_{x^\prime}^l\}_{l=1}^n$ where $T_{(\cdot)}^l \in \mathbb{R}^{h \times w \times d} $ and $x^\prime \in \mathcal{P}^\prime$, with $h$, $w$, and $d$ be the height, width, and dimension of the feature map $T$ respectively. At each layer $l$, the patch-level in-context residuals are captured by distances between the embeddings of the query token and the image prompt token across all image prompts in $\mathcal{P}^\prime$. Formally, for the query image $x$, its multi-layer patch-level in-context residuals at layer $l$ are modeled by a residual map $\mathbf{M}_x^l \in \mathbb{R}^{h \times w}$, where the residual value of each patch of $x$ is calculated based on its patch embedding and the nearest patch embedding of all images in $\mathcal{P}^\prime$ as:

\begin{equation}
    \mathbf{M}_x^l(i, j) = 1 - \langle T_x^l(i, j), h(T_x^l(i, j)|\mathcal{P}^\prime)\rangle,
    \label{eq:1}
\end{equation}
where $h(T_x^l(i, j)|\mathcal{P}^\prime)$ returns the embedding of the patch token that is most similar to $T_x^l(i, j)$ among all image patches in $\mathcal{P}^\prime$, and $\langle \cdot\rangle$ is the cosine similarity function. The final patch-level residual map $\mathbf{M}_x \in \mathbb{R}^{h \times w}$ is averaged over $n$ layer-wise residual maps:
\begin{equation}
    \mathbf{M}_{x} = \frac{1}{n} \sum_{l=1}^n \mathbf{M}_x^l.
    \label{eq:2}
\end{equation}

Each residual value in $\mathbf{M}_{x}$ is similar to a nearest-neighbor-distance anomaly score of the query patch to the image patch set in $\mathcal{P}^\prime$.  As shown in prior studies~\cite{cohen2020sub,defard2021padim,roth2022towards,pang2015lesinn,pang2018learning}, such distance-based anomaly scores can effectively discriminate anomalies from normal samples. Thus, the resulting residual map $\mathbf{M}_{x}$ provides a feature set of collective anomaly-discriminative power at multi-layer resolutions for the subsequent anomaly score learning in \coolname.

\subsection{Image-level Residual Learning}\label{subsec:image}
In addition to the discriminative power at the local patch-level residuals, the global discriminative information at the image level is also significant and serve as complementary knowledge to the patch-level features.

Hence, we introduce an image-level residual learning component to capture the higher-level discrepancies between $x$ and $\mathcal{P}^\prime$. Intuitively, the class token embedding from the last block of the visual encoder is used as the feature input, as it captures the most image-level discriminative information due to the bottom-up abstraction of information in the visual encoder. However, it is important to note that CLIP was originally designed for classification tasks, focusing on the semantic of the objects in the scenery,  which does not align well with the anomaly detection task in which both normal and abnormal samples are often from the same class of object. To reconcile this, we include an adapter layer $\psi(\cdot)$, parameterized by $\Theta_\psi$, to adapt the image representations further to anomaly detection, and thus, we learn the image-level residuals based on the adapted image features. Further, the prototypical features of the few-shot sample prompts, rather than the features of individual sample, are used to learn the in-context residuals, since they help capture more representative features of normal patterns.

Specifically, let $f_v(x)\in \mathbb{R}^{d^\prime}$ be the class token embedding of input $x$ in the visual encoder, we first compute the prototype of the feature maps of the image prompts in $\mathcal{P}^\prime$:

\begin{equation}
    \mathbf{I}_{p} = \frac{1}{K}\sum_{x_{k}^\prime \in \mathcal{P}^\prime}\psi(f_v(x_{k}^\prime);\Theta_\psi),
    \label{eq:3}
\end{equation}
where $\mathbf{I}_{p} \in \mathbb{R}^{d^\prime}$. Then let $\mathbf{I}_{x} = \psi(f_{v}(x);\Theta_\psi)$ be the adapted features of the query image $x$, the in-context image-level residual features $\mathbf{F}_{x}$ for $x$ are obtained  by performing element-wise subtraction between two feature maps:

\begin{equation}
    \mathbf{F}_{x} = \mathbf{I}_{x} \ominus \mathbf{I}_{p},
    \label{eq:4}
\end{equation}
where $\ominus$ denotes element-wise subtraction. Subsequently, these in-context residual features are fed to an image-level anomaly classification learner $\eta: \mathbf{F}_x \rightarrow \mathbb{R}$, parameterized by $\Theta_\eta$ which is optimized by the binary classification loss:
\begin{equation}
    \mathcal{L}_{IRL} = \frac{1}{N}\sum_{x\in X_{train}} \mathcal{L}_{b}(\eta(\mathbf{F}_{x};\Theta_\eta), y_x),
    \label{eq:5}
\end{equation}
where $\mathcal{L}_b$ is a binary classification loss. Focal loss \cite{lin2017focal} is used by default in our model.

\subsection{Fusing Text Prompt-based Prior Knowledge}\label{subsec:text}
The above two components are focused on residual learning based on the visual encoder. \coolname also allows easy incorporation of text-prompt-guided prior knowledge about normality and abnormality from the text encoder of CLIP. This helps \coolname leverage the normal and abnormal semantics hidden in the CLIP's pre-trained image-text-aligned embedding space for GAD. Motivated by this, \coolname exploits the text encoder to extract text prompt-guided discriminative features. Since the text prompts designed in WinCLIP~\cite{jeong2023winclip} show remarkable detection performance, \coolname adopts the same text prompt templates and its ensemble strategy, including both state and template-level text prompts. At the state level, generic text descriptions are employed to differentiate between normal and abnormal objects, whereas the template level provides a list of specific prompts tailored for anomaly detection (see \texttt{Appendix B.3} for detailed text prompts used). 

It should be noted that, unlike WinCLIP that uses these text prompts to directly compute the anomaly score, \coolname utilizes them to extract text-prompt-guided features for complementing the patch- and image-level residual features obtained through the visual encoder.

Specifically, let $\mathcal{P}^n_t$ be the set of text prompts for the normal class, we use the prototype of the text prompt embeddings to provide a representative embedding of the normal text prompts $\mathbf{F}_n=\frac{1}{|\mathcal{P}^n_t|}\sum_{p_i \in \mathcal{P}^n_t}f_t(p_i)$ where $p_i \in \mathcal{R}^{d^\prime}$; similarly we can obtain the prototype embedding for the abnormality text prompt set $\mathcal{P}^a_t$ by $\mathbf{F}_a=\frac{1}{|\mathcal{P}^a_t|}\sum_{p_j \in \mathcal{P}^a_t}f_t(p_j)$. Then, \coolname extracts an AD-oriented discriminative feature based on the similarity between the query image $x$ and the two prototypes of the text prompts:

\begin{equation}
    s_a(x) = \frac{\exp(\mathbf{F}_a^{\intercal}f_v(x))}{\exp(\mathbf{F}_n^{\intercal}f_v(x)) + \exp(\mathbf{F}_a^{\intercal}f_v(x))},
    \label{eq:6}
\end{equation}
where $[\cdot]^\intercal$ denotes a transpose operation, and $s_a(x)$ is the probability of the input $x$ being classified as abnormal.

\begin{table*}[ht]
\centering
\resizebox{0.92\textwidth}{!}{
\begin{tabular}{cc|ccccc|cc|cccc}
\hline
\multicolumn{1}{c|}{}                                         &                                    & \multicolumn{5}{c|}{}                                                                                                                                                                                                               & \multicolumn{2}{c|}{}                                                                     & \multicolumn{4}{c}{\textbf{Semantic Anomalies}}                                                                                                                                        \\ \cline{10-13} 
\multicolumn{1}{c|}{}                                         &                                    & \multicolumn{5}{c|}{\multirow{-2}{*}{\textbf{Industrial Defects}}}                                                                                                                                                                  & \multicolumn{2}{c|}{\multirow{-2}{*}{\textbf{Medical Anomalies}}}                         & \multicolumn{2}{c|}{\textbf{One-vs-all}}                                                   & \multicolumn{2}{c}{\textbf{Multi-class}}                                                  \\ \cline{3-13} 
\multicolumn{1}{c|}{\multirow{-3}{*}{\textbf{Setup}}} & \multirow{-3}{*}{\textbf{Methods}} & \textbf{ELPV}                               & \textbf{SDD}                                & \textbf{AITEX}                              & \textbf{VisA}                        & \textbf{MVTec AD}                     & \textbf{BrainMRI}                           & \textbf{HeadCT}                             & \multicolumn{1}{l}{\textbf{MNIST}}   & \multicolumn{1}{l|}{\textbf{CIFAR-10}} & \multicolumn{1}{l}{\textbf{MNIST}}   & \multicolumn{1}{l}{\textbf{CIFAR-10}} \\ \hline
\multicolumn{2}{c|}{\textbf{Baseline (0-shot)}}                                                    & 0.733{\scriptsize±0.000}                                 & 0.946{\scriptsize±0.000}                                 & 0.733{\scriptsize±0.000}                                 & 0.781{\scriptsize±0.000}                                 & 0.912{\scriptsize±0.000}                                 & 0.926{\scriptsize±0.000}                                 & 0.900{\scriptsize±0.000}                                 & 0.678{\scriptsize±0.000}                                 & 0.924{\scriptsize±0.000}                                  & 0.620{\scriptsize±0.000}                                 & 0.900{\scriptsize±0.000}                                 \\ \hline
\multicolumn{1}{c|}{}                                         & \textbf{SPADE}                     & 0.517{\scriptsize±0.012}                                 & 0.729{\scriptsize±0.041}                                 & 0.727{\scriptsize±0.004}                                 & 0.795{\scriptsize±0.045}                                 & 0.817{\scriptsize±0.054}                                 & 0.754{\scriptsize±0.048}                                 & 0.645{\scriptsize±0.034}                                 & 0.779{\scriptsize±0.024}                                 & 0.823{\scriptsize±0.014}                                  & 0.595{\scriptsize±0.060}                                 & 0.655{\scriptsize±0.042}                                 \\
\multicolumn{1}{c|}{}                                         & \textbf{PaDiM}                     & 0.594{\scriptsize±0.083}                                 & 0.721{\scriptsize±0.015}                                 & {\color[HTML]{FF0000} \textbf{0.784{\scriptsize±0.028}}} & 0.680{\scriptsize±0.042}                                 & 0.785{\scriptsize±0.025}                                 & 0.657{\scriptsize±0.122}                                 & 0.595{\scriptsize±0.036}                                & -                                 & -                                  & -                                 & -                               \\
\multicolumn{1}{c|}{}                                         & \textbf{Patchcore}                 & 0.716{\scriptsize±0.031}                                 & 0.902{\scriptsize±0.006}                                 & 0.739{\scriptsize±0.017}                                 & 0.817{\scriptsize±0.028}                                 & 0.858{\scriptsize±0.034}                                 & 0.706{\scriptsize±0.009}                                 & 0.736{\scriptsize±0.096}                                 &  0.756{\scriptsize±0.004}                                                                            & 0.602{\scriptsize±0.009}                                           & 0.603{\scriptsize±0.009}                                           & 0.703{\scriptsize±0.008}                                         \\
\multicolumn{1}{c|}{}                                         & \textbf{RegAD}                     & 0.571{\scriptsize±0.016}                                 & 0.499{\scriptsize±0.008}                                 & 0.564{\scriptsize±0.072}                                 & 0.557{\scriptsize±0.053}                                 & 0.640{\scriptsize±0.047}                                 & 0.449{\scriptsize±0.129}                                 & 0.602{\scriptsize±0.018}                                 & 0.525{\scriptsize±0.030}                                 & 0.534{\scriptsize±0.005}                                  & 0.608{\scriptsize±0.026}                                 & 0.695{\scriptsize±0.002}                                 \\
\multicolumn{1}{c|}{}                                         & \textbf{CoOp}                   & 0.762{\scriptsize±0.011}                                 & 0.897{\scriptsize±0.006}                                 & 0.687{\scriptsize±0.062}                                 & 0.806{\scriptsize±0.023}                                 & 0.858{\scriptsize±0.016}                                 & 0.725{\scriptsize±0.020}                                 & 0.811{\scriptsize±0.003}                                 & 0.557{\scriptsize±0.006}                                 & 0.527{\scriptsize±0.011}                                  & 0.612{\scriptsize±0.007}                                 & 0.393{\scriptsize±0.009}                                 \\
\multicolumn{1}{c|}{}                                         & \textbf{WinCLIP}                   & {\color[HTML]{0000FF} \textbf{0.726{\scriptsize±0.020}}} & {\color[HTML]{0000FF} \textbf{0.942{\scriptsize±0.006}}} & 0.726{\scriptsize±0.055}                                 & {\color[HTML]{0000FF} \textbf{0.842{\scriptsize±0.024}}} & {\color[HTML]{0000FF} \textbf{0.931{\scriptsize±0.019}}} & {\color[HTML]{0000FF} \textbf{0.934{\scriptsize±0.012}}} & {\color[HTML]{0000FF} \textbf{0.915{\scriptsize±0.015}}} & {\color[HTML]{0000FF} \textbf{0.810{\scriptsize±0.008}}} & {\color[HTML]{0000FF} \textbf{0.925{\scriptsize±0.001}}}  & {\color[HTML]{0000FF} \textbf{0.632{\scriptsize±0.000}}} & {\color[HTML]{0000FF} \textbf{0.914{\scriptsize±0.005}}} \\
\multicolumn{1}{c|}{\multirow{-7}{*}{\textbf{2-shot}}}       & \textbf{Ours (\coolname)}                      & {\color[HTML]{FF0000} \textbf{0.839{\scriptsize±0.003}}} & {\color[HTML]{FF0000} \textbf{0.972{\scriptsize±0.011}}} & {\color[HTML]{0000FF} \textbf{0.761{\scriptsize±0.029}}} & {\color[HTML]{FF0000} \textbf{0.858{\scriptsize±0.022}}} & {\color[HTML]{FF0000} \textbf{0.940{\scriptsize±0.015}}} & {\color[HTML]{FF0000} \textbf{0.973{\scriptsize±0.027}}} & {\color[HTML]{FF0000} \textbf{0.929{\scriptsize±0.025}}} & {\color[HTML]{FF0000} \textbf{0.892{\scriptsize±0.009}}} & {\color[HTML]{FF0000} \textbf{0.935{\scriptsize±0.002}}}  & {\color[HTML]{FF0000} \textbf{0.635{\scriptsize±0.010}}} & {\color[HTML]{FF0000} \textbf{0.924{\scriptsize±0.005}}} \\ \hline
\multicolumn{1}{c|}{}                                         & \textbf{SPADE}                     & 0.537{\scriptsize±0.013}                                 & 0.731{\scriptsize±0.020}                                 & 0.718{\scriptsize±0.011}                                 & 0.811{\scriptsize±0.040}                                 & 0.828{\scriptsize±0.044}                                 & 0.759{\scriptsize±0.070}                                 & 0.624{\scriptsize±0.012}                                 & 0.810{\scriptsize±0.009}                                 & 0.836{\scriptsize±0.006}                                  & 0.588{\scriptsize±0.041}                                 & 0.631{\scriptsize±0.063}                                 \\
\multicolumn{1}{c|}{}                                         & \textbf{PaDiM}                     & 0.612{\scriptsize±0.080}                                 & 0.742{\scriptsize±0.014}                                 & {\color[HTML]{0000FF} \textbf{0.787{\scriptsize±0.038}}} & 0.735{\scriptsize±0.031}                                 & 0.805{\scriptsize±0.018}                                 & 0.792{\scriptsize±0.048}                                 & 0.622{\scriptsize±0.013}                                 & -                                & -                                  & -                                & -                                 \\
\multicolumn{1}{c|}{}                                         & \textbf{Patchcore}                 & 0.756{\scriptsize±0.073} & 0.923{\scriptsize±0.008}                                 & 0.733{\scriptsize±0.002}                                 & 0.843{\scriptsize±0.025}                                 & 0.885{\scriptsize±0.026}                                 & 0.794{\scriptsize±0.040}                                 & 0.805{\scriptsize±0.006}                                  & 0.833{\scriptsize±0.009}                                 & 0.639{\scriptsize±0.010}                                  & 0.497{\scriptsize±0.044}                                 & 0.739{\scriptsize±0.011}                                           \\
\multicolumn{1}{c|}{}                                         & \textbf{RegAD}                     & 0.596{\scriptsize±0.040}                                 & 0.525{\scriptsize±0.027}                                 & 0.596{\scriptsize±0.074}                                 & 0.574{\scriptsize±0.042}                                 & 0.663{\scriptsize±0.032}                                 & 0.571{\scriptsize±0.149}                                 & 0.522{\scriptsize±0.050}                                 & 0.548{\scriptsize±0.053}                                 & 0.534{\scriptsize±0.002}                                  & 0.596{\scriptsize±0.075}                                 & 0.677{\scriptsize±0.161}                                 \\
\multicolumn{1}{c|}{}                                         & \textbf{CoOp}                   & {\color[HTML]{0000FF} \textbf{0.781{\scriptsize±0.002}}}                                 & 0.902{\scriptsize±0.006}                                 & 0.720{\scriptsize±0.017}                                 & 0.818{\scriptsize±0.018}                                & 0.874{\scriptsize±0.017}                                 & 0.759{\scriptsize±0.033}                                 & 0.860{\scriptsize±0.032}                                 & 0.563{\scriptsize±0.004}                                 & 0.537{\scriptsize±0.005}                                  & 0.618{\scriptsize±0.002}                                 & 0.395{\scriptsize±0.008}                                 \\
\multicolumn{1}{c|}{}                                         & \textbf{WinCLIP}                   & 0.754{\scriptsize±0.009}                                 & {\color[HTML]{0000FF} \textbf{0.943{\scriptsize±0.004}}} & 0.764{\scriptsize±0.025}                                 & {\color[HTML]{0000FF} \textbf{0.858{\scriptsize±0.025}}} & {\color[HTML]{0000FF} \textbf{0.940{\scriptsize±0.021}}} & {\color[HTML]{0000FF} \textbf{0.941{\scriptsize±0.002}}} & {\color[HTML]{0000FF} \textbf{0.912{\scriptsize±0.003}}} & {\color[HTML]{0000FF} \textbf{0.851{\scriptsize±0.010}}} & {\color[HTML]{0000FF} \textbf{0.927{\scriptsize±0.001}}}  & {\color[HTML]{0000FF} \textbf{0.632{\scriptsize±0.004}}} & {\color[HTML]{0000FF} \textbf{0.915{\scriptsize±0.003}}} \\
\multicolumn{1}{c|}{\multirow{-7}{*}{\textbf{4-shot}}}       & \textbf{Ours (\coolname)}                      & {\color[HTML]{FF0000} \textbf{0.846{\scriptsize±0.011}}} & {\color[HTML]{FF0000} \textbf{0.975{\scriptsize±0.006}}} & {\color[HTML]{FF0000} \textbf{0.790{\scriptsize±0.018}}} & {\color[HTML]{FF0000} \textbf{0.877{\scriptsize±0.019}}} & {\color[HTML]{FF0000} \textbf{0.945{\scriptsize±0.018}}} & {\color[HTML]{FF0000} \textbf{0.975{\scriptsize±0.016}}} & {\color[HTML]{FF0000} \textbf{0.933{\scriptsize±0.013}}} & {\color[HTML]{FF0000} \textbf{0.902{\scriptsize±0.016}}} & {\color[HTML]{FF0000} \textbf{0.940{\scriptsize±0.010}}}  & {\color[HTML]{FF0000} \textbf{0.643{\scriptsize±0.007}}} & {\color[HTML]{FF0000} \textbf{0.928{\scriptsize±0.009}}} \\ \hline
\multicolumn{1}{c|}{}                                         & \textbf{SPADE}                     & 0.567{\scriptsize±0.034}                                 & 0.741{\scriptsize±0.011 }                                & 0.708{\scriptsize±0.006}                                 & 0.821{\scriptsize±0.042}                                 & 0.840{\scriptsize±0.057 }                                & 0.794{\scriptsize±0.039}                                 & 0.626{\scriptsize±0.022}                                 & 0.829{\scriptsize±0.009 }                                & 0.849{\scriptsize±0.006}                                  & 0.597{\scriptsize±0.028}                                 & 0.656{\scriptsize±0.037}                                 \\
\multicolumn{1}{c|}{}                                         & \textbf{PaDiM}                     & 0.724{\scriptsize±0.017}                                 & 0.769{\scriptsize±0.037}                                 & 0.792{\scriptsize±0.025}                                 & 0.768{\scriptsize±0.032}                                 & 0.820{\scriptsize±0.016}                                 & 0.758{\scriptsize±0.025 }                                & 0.661{\scriptsize±0.039}                                 & - & -                                 & -                              & -                               \\
\multicolumn{1}{c|}{}                                         & \textbf{Patchcore}                 & {\color[HTML]{0000FF} \textbf{0.837{\scriptsize±0.016}}} & 0.925{\scriptsize±0.003}                                 & 0.745{\scriptsize±0.002}                                 & 0.860{\scriptsize±0.026}                                 & 0.922{\scriptsize±0.019}                                 & 0.812{\scriptsize±0.016 }                                & 0.817{\scriptsize±0.034}                                 &{\color[HTML]{0000FF} \textbf{0.876{\scriptsize±0.004}}} & 0.672{\scriptsize±0.006}                                  & 0.526{\scriptsize±0.019}                                 & 0.764{\scriptsize±0.004}                                          \\
\multicolumn{1}{c|}{}                                         & \textbf{RegAD}                     & 0.633{\scriptsize±0.027 }                                & 0.594{\scriptsize±0.029}                                 & 0.603{\scriptsize±0.062}                                 & 0.589{\scriptsize±0.040}                                 & 0.674{\scriptsize±0.033 }                                & 0.632{\scriptsize±0.079}                                 & 0.628{\scriptsize±0.026 }                                & 0.547{\scriptsize±0.063 }                                & 0.555{\scriptsize±0.008 }                                 & 0.573{\scriptsize±0.076}                                 & 0.587{\scriptsize±0.211}                                 \\
\multicolumn{1}{c|}{}                                         & \textbf{CoOp}                   & 0.817{\scriptsize±0.012}                                & 0.898{\scriptsize±0.005 }                                & 0.769{\scriptsize±0.008 }                                & 0.822{\scriptsize±0.021}                                 & 0.880{\scriptsize±0.014}                                 & 0.755{\scriptsize±0.003}                                 & 0.914{\scriptsize±0.027}                                 & 0.567{\scriptsize±0.007}                                 & 0.542{\scriptsize±0.005}                                  & 0.619{\scriptsize±0.004}                                 & 0.399{\scriptsize±0.006 }                                \\
\multicolumn{1}{c|}{}                                         & \textbf{WinCLIP}                   & 0.814{\scriptsize±0.010}                                 & {\color[HTML]{0000FF} \textbf{0.941{\scriptsize±0.001}}} & {\color[HTML]{0000FF} \textbf{0.796{\scriptsize±0.015}}} & {\color[HTML]{0000FF} \textbf{0.868{\scriptsize±0.020}}} & {\color[HTML]{0000FF} \textbf{0.947{\scriptsize±0.025}}} & {\color[HTML]{0000FF} \textbf{0.944{\scriptsize±0.001}}} & {\color[HTML]{0000FF} \textbf{0.915{\scriptsize±0.008}}} & 0.867{\scriptsize±0.007}                                 & {\color[HTML]{0000FF} \textbf{0.928{\scriptsize±0.001}}}  & {\color[HTML]{0000FF} \textbf{0.641{\scriptsize±0.004}}} & {\color[HTML]{0000FF} \textbf{0.916{\scriptsize±0.003}}} \\
\multicolumn{1}{c|}{\multirow{-7}{*}{\textbf{8-shot}}}       & \textbf{Ours (\coolname)}                      & {\color[HTML]{FF0000} \textbf{0.872{\scriptsize±0.013}}} & {\color[HTML]{FF0000} \textbf{0.978{\scriptsize±0.006}}} & {\color[HTML]{FF0000} \textbf{0.806{\scriptsize±0.036}}} & {\color[HTML]{FF0000} \textbf{0.887{\scriptsize±0.021}}} & {\color[HTML]{FF0000} \textbf{0.953{\scriptsize±0.013}}} & {\color[HTML]{FF0000} \textbf{0.983{\scriptsize±0.012}}} & {\color[HTML]{FF0000} \textbf{0.936{\scriptsize±0.008}}} & {\color[HTML]{FF0000} \textbf{0.920{\scriptsize±0.003}}}  & {\color[HTML]{FF0000} \textbf{0.945{\scriptsize±0.002}}}  & {\color[HTML]{FF0000} \textbf{0.646{\scriptsize±0.003}}} & {\color[HTML]{FF0000} \textbf{0.934{\scriptsize±0.008}}} \\ \hline
\end{tabular}
}
\caption{AUROC results(mean±std) on nine real-world AD datasets under various few-shot AD settings. Best results and the second-best results are respectively highlighted in {\color[HTML]{FF0000} \textbf{red}} and {\color[HTML]{0000FF} \textbf{blue}}. `Baseline' is a WinCLIP-based zero-shot AD model. }
\label{auroc}
\vspace{-0.5cm}
\end{table*}

\subsection{Training and Inference}\label{subsec:train_inference}
\vspace{0.1cm}
\noindent\textbf{In-Context Residual Learning.} During training, \coolname performs a holistic residual learning that synthesizes both patch-level and image-level residual information, augmented by the text prompt-guided features. The holistic in-context residual map of a query image $x$ is defined as:

\begin{equation}
    \mathbf{M}_{x}^{+} = \mathbf{M}_{x} \oplus s_i(x) \oplus s_a(x),
    \label{eq:7}
\end{equation}
where $s_{i}(x)=\eta(\mathbf{F}_x;\Theta_\eta)$ is an anomaly score based on the image-level residual map $\mathbf{F}_x$ and $\oplus$ denotes an element-wise addition. \coolname then devises a holistic anomaly scoring function $\phi$, parameterized by $\Theta_\phi$, based on $\mathbf{M}_{x}^{+}$, and defines the final anomaly score as:

\begin{equation}
    s(x) = \phi(\mathbf{M}_{x}^{+};\Theta_\phi) + \alpha s_p(x),
    \label{eq:finalscore}
\end{equation}
where $\phi(\mathbf{M}_{x}^{+};\Theta_\phi)$ performs a holistic anomaly scoring using patch-, image-level and text prompt-guided features, while $s_p(x)=\max(\mathbf{M}_{x})$ is a maximum residual score-based fine-grained anomaly score at the image patch level. $s_p(x)$ is added into Eq. \ref{eq:finalscore} because such patch-level anomaly scores are crucial for detecting local abnormal regions to which the $\phi$-based holistic anomaly score can often overlook. $\alpha$ is a hyper-parameter that modulates the contribution of the patch-level residual score. Lastly, we optimize the final anomaly score $s(x)$ using $X_{train}$:
\begin{equation}
    \mathcal{L}_{h}=\frac{1}{N}\sum_{x \in X_{train}} \mathcal{L}_b(s(x),y_x).
    \label{eq:8}
\end{equation}

Thus, the full \coolname model is optimized by minimizing the overall loss as follows:
\begin{equation}
    \mathcal{L}_{\mathit{InCTRL}} = \mathcal{L}_{IRL} + \mathcal{L}_{h}.
    \label{eq:overall_loss}
\end{equation}

\vspace{0.1cm}
\noindent\textbf{Inference.}
During inference, for a given test image $x_t$ and the $K$-shot normal image prompt set $\mathcal{P}$ from the target dataset, they are fed forward through the visual encoder and the adapter layers, obtaining $\mathbf{M}_{x_t}$ and $s_i(x_t)$. The text prompt sets used during training are used to obtain $s_a(x_t)$. Lastly, we obtain the final anomaly score of $x_t$ via Eq. \ref{eq:finalscore}.

\section{Experiments}
\subsection{Experimental Setup}

\vspace{0.1cm}
\noindent\textbf{Datasets.} 
To verify the efficiency of our method \coolname, we conduct comprehensive experiments across nine real-world AD datasets, including five industrial defect inspection dataset (MVTec AD~\cite{bergmann2019mvtec}, VisA~\cite{zou2022spot}, AITEX~\cite{silvestre2019public}, ELPV~\cite{deitsch2019automatic}, SDD~\cite{tabernik2020segmentation}), two medical image datasets (BrainMRI~\cite{salehi2021multiresolution}, HeadCT~\cite{salehi2021multiresolution}), and two semantic anomaly detection datasets: MNIST~\cite{lecun1998gradient} and CIFAR-10~\cite{krizhevsky2009cifar} under both one-vs-all and multi-class protocols~\cite{Cao_2023_ICCV,ruff2020deep}. Under the one-vs-all protocol, one class is used as normal, with the other classes treated as abnormal; while under the multi-class protocol, images of even-number classes from MNIST and animal-related classes from CIFAR-10 are treated as normal, with the images of the other classes are considered as anomalies (see \texttt{Appendix A} for more details).

To assess the GAD performance, MVTec AD, the combination of its training and test sets, is used as the auxiliary training data, on which GAD models are trained, and they are subsequently evaluated on the test set of the other eight datasets without any further training. We train the model on VisA when evaluating the performance on MVTec AD. 
The few-shot normal prompts for the target data are randomly sampled from the training set of target datasets and remain the same for all models for fair comparison. We evaluate the performance with the number of few-shot normal prompt set to $K = 2, 4, 8$. The reported results are averaged over three independent runs with different random seeds. 

\vspace{0.1cm}
\noindent\textbf{Competing Methods and Evaluation Metrics.}
Since we aim to achieve a generalist AD model, the comparision is focus on detectors of similar generalist detection capabilities. Following \cite{jeong2023winclip}, \coolname is compared with three conventional full-shot AD approaches, including SPADE~\cite{cohen2020sub}, PaDiM~\cite{defard2021padim}, and PatchCore~\cite{roth2022towards}, all of which are adapted to the few-shot setting by performing their distance-based anomaly scoring based on the few-shot normal samples. We also compare with state-of-the-art (SotA) conventional few-shot AD method RegAD~\cite{huang2022registration} and the CLIP-driven method WinCLIP~\cite{jeong2023winclip}. The popular prompt learning method CoOp~\cite{zhou2022conditional} is used as an additional baseline that is trained on the auxiliary data as \coolname, after which it uses the few-shot anomaly scoring strategy in WinCLIP to perform anomaly detection. 

As for evaluation metrics, following previous works~\cite{jeong2023winclip,huang2022registration,roth2022towards,Cao_2023_ICCV,pang2023deep}, we use two popular metrics AUROC (Area Under the Receiver Operating Characteristic) and AUPRC (Area Under the Precision-Recall Curve) to evaluate the AD performance. We also evaluate the number of parameters and per-image inference time of CLIP-based methods, which is presented in our \texttt{Appendix C.1}. 

\begin{table*}[ht]
\centering
\resizebox{0.92\textwidth}{!}{
\begin{tabular}{cc|ccccc|cc|cccc}
\hline
\multicolumn{1}{c|}{}                                         &                                    & \multicolumn{5}{c|}{}                                                                                                                                                                                                               & \multicolumn{2}{c|}{}                                                                     & \multicolumn{4}{c}{\textbf{Semantic Anomalies}}                                                                                                                                        \\ \cline{10-13} 
\multicolumn{1}{c|}{}                                         &                                    & \multicolumn{5}{c|}{\multirow{-2}{*}{\textbf{Industrial Defects}}}                                                                                                                                                                  & \multicolumn{2}{c|}{\multirow{-2}{*}{\textbf{Medical Anomalies}}}                         & \multicolumn{2}{c|}{\textbf{One-vs-all}}                                                   & \multicolumn{2}{c}{\textbf{Multi-class}}                                                  \\ \cline{3-13} 
\multicolumn{1}{c|}{\multirow{-3}{*}{\textbf{Setup}}} & \multirow{-3}{*}{\textbf{Methods}} & \textbf{ELPV}                               & \textbf{SDD}                                & \textbf{AITEX}                              & \textbf{VisA}                        & \textbf{MVTec AD}                     & \textbf{BrainMRI}                           & \textbf{HeadCT}                             & \multicolumn{1}{l}{\textbf{MNIST}}   & \multicolumn{1}{l|}{\textbf{CIFAR-10}} & \multicolumn{1}{l}{\textbf{MNIST}}   & \multicolumn{1}{l}{\textbf{CIFAR-10}} \\ \hline
\multicolumn{2}{c|}{\textbf{Baseline (0-shot)}}                                                    & 0.855{\scriptsize±0.000}                        & 0.886{\scriptsize±0.000}                       & 0.552{\scriptsize±0.000}                       & 0.812{\scriptsize±0.000}                       & 0.957{\scriptsize±0.000}                        & 0.988{\scriptsize±0.000}                        & 0.970{\scriptsize±0.000}                       & 0.940{\scriptsize±0.000}                                 & 0.990{\scriptsize±0.000}                                  & 0.606{\scriptsize±0.000}                                 & 0.852{\scriptsize±0.000}                                 \\ \hline
\multicolumn{1}{c|}{}                                         & \textbf{SPADE}                     & 0.618{\scriptsize±0.007}                                 & 0.366{\scriptsize±0.105}                                 & 0.470{\scriptsize±0.008}                                 & 0.818{\scriptsize±0.031}                                 & 0.922{\scriptsize±0.023}                                 & 0.952{\scriptsize±0.009}                                 & 0.851{\scriptsize±0.022}                                 & {\color[HTML]{0000FF} \textbf{0.965{\scriptsize±0.004}}} & 0.971{\scriptsize±0.003}                                  & {\color[HTML]{0000FF} \textbf{0.615{\scriptsize±0.068}}} & 0.502{\scriptsize±0.035 }                                \\
\multicolumn{1}{c|}{}                                         & \textbf{PaDiM}                     & 0.707{\scriptsize±0.058}                                 & 0.337{\scriptsize±0.008}                                 & {\color[HTML]{FF0000} \textbf{0.529{\scriptsize±0.034}}} & 0.719{\scriptsize±0.027}                                 & 0.890{\scriptsize±0.015}                                 & 0.902{\scriptsize±0.046}                                 & 0.876{\scriptsize±0.017}                                 & -                                 & -                                  & -                                & -                             \\
\multicolumn{1}{c|}{}                                         & \textbf{Patchcore}                 & 0.840{\scriptsize±0.031}                                 & 0.676{\scriptsize±0.003}                                 & 0.378{\scriptsize±0.008}                                 & 0.841{\scriptsize±0.023}                                 & 0.939{\scriptsize±0.012}                                 & 0.921{\scriptsize±0.017}                                 & 0.913{\scriptsize±0.002}                                 & 0.956{\scriptsize±0.001}                                 & 0.926{\scriptsize±0.002}                                  & 0.482{\scriptsize±0.025}                                 & 0.574{\scriptsize±0.015}                                   \\
\multicolumn{1}{c|}{}                                         & \textbf{RegAD}                     & 0.679{\scriptsize±0.005}                                 & 0.173{\scriptsize±0.019}                                 & 0.275{\scriptsize±0.035}                                 & 0.614{\scriptsize±0.037}                                 & 0.837{\scriptsize±0.034}                                 & 0.872{\scriptsize±0.065}                                 & 0.854{\scriptsize±0.009}                                 & 0.913{\scriptsize±0.006}                                 & 0.909{\scriptsize±0.003}                                  & 0.612{\scriptsize±0.013}                                 & 0.672{\scriptsize±0.008}                                 \\
\multicolumn{1}{c|}{}                                         & \textbf{CoOp}                   & 0.841{\scriptsize±0.020}                                 & 0.543{\scriptsize±0.004}                                 & 0.443{\scriptsize±0.050}                                 & 0.835{\scriptsize±0.019}                                 & 0.922{\scriptsize±0.007}                                 & 0.923{\scriptsize±0.002}                                 & 0.937{\scriptsize±0.014}                                 & 0.926{\scriptsize±0.003}                                 & 0.911{\scriptsize±0.002}                                  & 0.607{\scriptsize±0.009}                                 & 0.371{\scriptsize±0.013}                                 \\
\multicolumn{1}{c|}{}                                         & \textbf{WinCLIP}                   & {\color[HTML]{0000FF} \textbf{0.849{\scriptsize±0.010}}} & {\color[HTML]{0000FF} \textbf{0.865{\scriptsize±0.004}}} & 0.500{\scriptsize±0.043}                                 & {\color[HTML]{0000FF} \textbf{0.859{\scriptsize±0.021}}} & {\color[HTML]{0000FF} \textbf{0.965{\scriptsize±0.007}}} & {\color[HTML]{0000FF} \textbf{0.989{\scriptsize±0.003}}} & {\color[HTML]{0000FF} \textbf{0.975{\scriptsize±0.012}}} & 0.963{\scriptsize±0.001}                                 & {\color[HTML]{0000FF} \textbf{0.990{\scriptsize±0.001}}}  & 0.614{\scriptsize±0.005}                                 & {\color[HTML]{0000FF} \textbf{0.876{\scriptsize±0.016}}} \\
\multicolumn{1}{c|}{\multirow{-7}{*}{\textbf{2-shot}}}       & \textbf{Ours (\coolname)}                      & {\color[HTML]{FF0000} \textbf{0.913{\scriptsize±0.008}}} & {\color[HTML]{FF0000} \textbf{0.917{\scriptsize±0.009}}} & {\color[HTML]{0000FF} \textbf{0.519{\scriptsize±0.022}}} & {\color[HTML]{FF0000} \textbf{0.877{\scriptsize±0.016}}} & {\color[HTML]{FF0000} \textbf{0.969{\scriptsize±0.004}}} & {\color[HTML]{FF0000} \textbf{0.994{\scriptsize±0.013}}} & {\color[HTML]{FF0000} \textbf{0.981{\scriptsize±0.013}}} & {\color[HTML]{FF0000} \textbf{0.975{\scriptsize±0.004}}} & {\color[HTML]{FF0000} \textbf{0.992{\scriptsize±0.000}}}  & {\color[HTML]{FF0000} \textbf{0.618{\scriptsize±0.012}}} & {\color[HTML]{FF0000} \textbf{0.899{\scriptsize±0.010}}} \\ \hline
\multicolumn{1}{c|}{}                                         & \textbf{SPADE}                     & 0.627{\scriptsize±0.011}                                 & 0.385{\scriptsize±0.018}                                 & 0.451{\scriptsize±0.031}                                 & 0.826{\scriptsize±0.024}                                 & 0.924{\scriptsize±0.015}                                 & 0.958{\scriptsize±0.017}                                 & 0.854{\scriptsize±0.016}                                 & 0.966{\scriptsize±0.008}                                 & 0.973{\scriptsize±0.002}                                  & {\color[HTML]{0000FF} \textbf{0.611{\scriptsize±0.053}}} & 0.487{\scriptsize±0.047}                                 \\
\multicolumn{1}{c|}{}                                         & \textbf{PaDiM}                     & 0.724{\scriptsize±0.067}                                 & 0.351{\scriptsize±0.012}                                 & {\color[HTML]{0000FF} \textbf{0.540{\scriptsize±0.053}}} & 0.758{\scriptsize±0.018 }                                & 0.909{\scriptsize±0.013}                                 & 0.956{\scriptsize±0.011}                                 & 0.890{\scriptsize±0.011}                                 & - & -                                & -                                 & -                               \\
\multicolumn{1}{c|}{}                                         & \textbf{Patchcore}                 & {\color[HTML]{0000FF} \textbf{0.871{\scriptsize±0.042}}} & 0.703{\scriptsize±0.013}                                 & 0.377{\scriptsize±0.001}                                 & 0.860{\scriptsize±0.016 }                                & 0.950{\scriptsize±0.013 }                                & 0.945{\scriptsize±0.017}                                 & 0.941{\scriptsize±0.009}                                  & {\color[HTML]{0000FF} \textbf{0.972{\scriptsize±0.002}}} & 0.934{\scriptsize±0.003}                                  & 0.504{\scriptsize±0.025}                                 & 0.606{\scriptsize±0.010 }                                          \\
\multicolumn{1}{c|}{}                                         & \textbf{RegAD}                     & 0.688{\scriptsize±0.018}                                 & 0.176{\scriptsize±0.003}                                 & 0.294{\scriptsize±0.031}                                 & 0.628{\scriptsize±0.034}                                 & 0.846{\scriptsize±0.026}                                 & 0.900{\scriptsize±0.041}                                 & 0.810{\scriptsize±0.028}                                 & 0.916{\scriptsize±0.013}                                 & 0.908{\scriptsize±0.001}                                  & 0.522{\scriptsize±0.085}                                 & 0.681{\scriptsize±0.127}                                 \\
\multicolumn{1}{c|}{}                                         & \textbf{CoOp}                   & 0.867{\scriptsize±0.003}                                 & 0.594{\scriptsize±0.014}                                 & 0.454{\scriptsize±0.014}                                 & 0.842{\scriptsize±0.016}                                 & 0.924{\scriptsize±0.008}                                 & 0.932{\scriptsize±0.013}                                 & 0.957{\scriptsize±0.017}                                 & 0.929{\scriptsize±0.002}                                 & 0.915{\scriptsize±0.003}                                  & 0.611{\scriptsize±0.003}                                 & 0.374{\scriptsize±0.012}                                 \\
\multicolumn{1}{c|}{}                                         & \textbf{WinCLIP}                   & 0.864{\scriptsize±0.004}                                 & {\color[HTML]{0000FF} \textbf{0.868{\scriptsize±0.003}}} & 0.513{\scriptsize±0.017}                                 & {\color[HTML]{0000FF} \textbf{0.875{\scriptsize±0.023}}} & {\color[HTML]{0000FF} \textbf{0.968{\scriptsize±0.008}}} & {\color[HTML]{0000FF} \textbf{0.990{\scriptsize±0.001}}} & {\color[HTML]{0000FF} \textbf{0.974{\scriptsize±0.002}}} & 0.971{\scriptsize±0.002}                                 & {\color[HTML]{0000FF} \textbf{0.990{\scriptsize±0.000}}}  & {\color[HTML]{0000FF} \textbf{0.611{\scriptsize±0.011}}} & {\color[HTML]{0000FF} \textbf{0.882{\scriptsize±0.009}}} \\
\multicolumn{1}{c|}{\multirow{-7}{*}{\textbf{4-shot}}}       & \textbf{Ours (\coolname)}                      & {\color[HTML]{FF0000} \textbf{0.916{\scriptsize±0.009}}} & {\color[HTML]{FF0000} \textbf{0.924{\scriptsize±0.015}}} & {\color[HTML]{FF0000} \textbf{0.548{\scriptsize±0.016}}} & {\color[HTML]{FF0000} \textbf{0.902{\scriptsize±0.027}}} & {\color[HTML]{FF0000} \textbf{0.972{\scriptsize±0.006}}} & {\color[HTML]{FF0000} \textbf{0.994{\scriptsize±0.013}}} & {\color[HTML]{FF0000} \textbf{0.984{\scriptsize±0.011}}} & {\color[HTML]{FF0000} \textbf{0.980{\scriptsize±0.007}}} & {\color[HTML]{FF0000} \textbf{0.992{\scriptsize±0.004}}}  & {\color[HTML]{FF0000} \textbf{0.620{\scriptsize±0.004}}} & {\color[HTML]{FF0000} \textbf{0.901{\scriptsize±0.020}}} \\ \hline
\multicolumn{1}{c|}{}                                         & \textbf{SPADE}                     & 0.641{\scriptsize±0.018}                                 & 0.394{\scriptsize±0.024}                                 & 0.427{\scriptsize±0.008}                                 & 0.844{\scriptsize±0.031}                                 & 0.930{\scriptsize±0.016}                                 & 0.962{\scriptsize±0.014}                                 & 0.860{\scriptsize±0.019}                                 & 0.974{\scriptsize±0.002}                                 & 0.976{\scriptsize±0.001}                                  & 0.613{\scriptsize±0.035}                                 & 0.515{\scriptsize±0.024}                                 \\
\multicolumn{1}{c|}{}                                         & \textbf{PaDiM}                     & 0.798{\scriptsize±0.014}                                 & 0.384{\scriptsize±0.045}                                 & 0.555{\scriptsize±0.031}                                 & 0.781{\scriptsize±0.024}                                 & 0.927{\scriptsize±0.012}                                 & 0.946{\scriptsize±0.007}                                 & 0.896{\scriptsize±0.009}                                 & - & -                                 & -                                & -                                \\
\multicolumn{1}{c|}{}                                         & \textbf{Patchcore}                 & {\color[HTML]{0000FF} \textbf{0.915{\scriptsize±0.007}}} & 0.708{\scriptsize±0.009}                                 & 0.389{\scriptsize±0.003}                                 & 0.873{\scriptsize±0.022}                                 & 0.962{\scriptsize±0.013}                                 & 0.957{\scriptsize±0.007}                                 & 0.931{\scriptsize±0.006}                                  & {\color[HTML]{0000FF} \textbf{0.979{\scriptsize±0.001}}} & 0.942{\scriptsize±0.002}                                  & 0.530{\scriptsize±0.037}                                 & 0.635{\scriptsize±0.019}                                          \\
\multicolumn{1}{c|}{}                                         & \textbf{RegAD}                     & 0.696{\scriptsize±0.015}                                 & 0.246{\scriptsize±0.031}                                 & 0.314{\scriptsize±0.036}                                 & 0.643{\scriptsize±0.032}                                 & 0.855{\scriptsize±0.021}                                 & 0.908{\scriptsize±0.013}                                 & 0.881{\scriptsize±0.014}                                 & 0.919{\scriptsize±0.018}                                 & 0.911{\scriptsize±0.001}                                  & 0.566{\scriptsize±0.048}                                 & 0.558{\scriptsize±0.159}                                 \\
\multicolumn{1}{c|}{}                                         & \textbf{CoOp}                   & 0.905{\scriptsize±0.008}                                 & 0.578{\scriptsize±0.001}                                 & 0.514{\scriptsize±0.003}                                 & 0.848{\scriptsize±0.020}                                 & 0.933{\scriptsize±0.007}                                 & 0.927{\scriptsize±0.007}                                 & 0.965{\scriptsize±0.018}                                 & 0.937{\scriptsize±0.004}                                 & 0.920{\scriptsize±0.003}                                  & 0.610{\scriptsize±0.001}                                 & 0.376{\scriptsize±0.003}                                 \\
\multicolumn{1}{c|}{}                                         & \textbf{WinCLIP}                   & 0.897{\scriptsize±0.007}                                 & {\color[HTML]{0000FF} \textbf{0.865{\scriptsize±0.001}}} & {\color[HTML]{0000FF} \textbf{0.562{\scriptsize±0.024}}} & {\color[HTML]{0000FF} \textbf{0.880{\scriptsize±0.021}}} & {\color[HTML]{0000FF} \textbf{0.973{\scriptsize±0.009}}} & {\color[HTML]{0000FF} \textbf{0.991{\scriptsize±0.000}}} & {\color[HTML]{0000FF} \textbf{0.975{\scriptsize±0.003}}} & 0.974{\scriptsize±0.001}                                 & {\color[HTML]{0000FF} \textbf{0.990{\scriptsize±0.000}}}  & {\color[HTML]{0000FF} \textbf{0.616{\scriptsize±0.006}}} & {\color[HTML]{0000FF} \textbf{0.887{\scriptsize±0.006}}} \\
\multicolumn{1}{c|}{\multirow{-7}{*}{\textbf{8-shot}}}       & \textbf{Ours (\coolname)}                      & {\color[HTML]{FF0000} \textbf{0.926{\scriptsize±0.006}}} & {\color[HTML]{FF0000} \textbf{0.925{\scriptsize±0.011}}} & {\color[HTML]{FF0000} \textbf{0.561{\scriptsize±0.034}}} & {\color[HTML]{FF0000} \textbf{0.904{\scriptsize±0.025}}} & {\color[HTML]{FF0000} \textbf{0.977{\scriptsize±0.006}}} & {\color[HTML]{FF0000} \textbf{0.996{\scriptsize±0.003}}} & {\color[HTML]{FF0000} \textbf{0.985{\scriptsize±0.005}}} & {\color[HTML]{FF0000} \textbf{0.989{\scriptsize±0.001}}} & {\color[HTML]{FF0000} \textbf{0.994{\scriptsize±0.001}}}  & {\color[HTML]{FF0000} \textbf{0.622{\scriptsize±0.008}}} & {\color[HTML]{FF0000} \textbf{0.912{\scriptsize±0.005}}} \\ \hline
\end{tabular}
}
\caption{AUPRC results(mean±std) on nine real-world AD datasets under various few-shot AD settings. Best results and the second-best results are respectively highlighted in {\color[HTML]{FF0000} \textbf{red}} and {\color[HTML]{0000FF} \textbf{blue}}. `Baseline' is a WinCLIP-based zero-shot AD model. }
\label{aupr}
\vspace{-0.3cm}
\end{table*}

\vspace{0.1cm}
\noindent\textbf{Implementation Details.}
By default, for CLIP-based models, including WinCLIP, CoOp and our \coolname, we adopt the same CLIP implementation, OpenCLIP~\cite{ilharco2021openclip}, and its public pre-trained backbone ViT-B/16+ in our experiments. Adam is used as the optimizer and the initial learning rate is set to 1e-3 by default. 
The text prompts used in \coolname are kept exactly the same as WinCLIP. To enable the model to recognize both normal and abnormal objects while preventing overfitting, the training epochs is set to 10 with a batch size of 48 on a single GPU (NVIDIA GeForce RTX 3090). 
SPADE, PaDiM and WinCLIP\footnote{No official implementation of WinCLIP is available. Our implementation is available at \url{https://github.com/mala-lab/WinCLIP}.} use the same image prompts as \coolname for fair comparison, and the official implementation of PatchCore, RegAD and CoOp is taken. Further details are provided in \texttt{Appendix B}.

\subsection{Main Results}
Tables~\ref{auroc} and~\ref{aupr} present the comparison results of \coolname to six SotA competing methods in AUROC and AUPRC, respectively, on nine real-world AD datasets.
Note that the results for MVTec AD, VisA, and the one-vs-all settings of MNIST and CIFAR-10 represent an average result across their respective data subsets (see \texttt{Appendix C} for breakdown results). Below we analyze these results in detail.

\vspace{0.1cm}
\noindent\textbf{Generalization to Industrial Defects.}
Generally, for the five industrial defect AD datasets, \coolname significantly outperforms all competing models on almost all cases across the three few-shot settings. With more few-shot image prompts, the performance of all methods generally gets better. Specifically, Patchcore shows better performance than SPADE, PaDiM and RegAD, but all of which generalize badly on these datasets. WinCLIP obtains fairly good generalization and surpasses Patchcore, owing to CLIP's superior recognition capabilities. Due to the in-context residual information is well transferable across the datasets, \coolname exhibits superior performance, outperforming the SotA models by a large margin, particularly on challenging datasets like ELPV and SDD. As a result, our \coolname model respectively gains up to 11.3\%, 6.5\%, 3.7\% AUROC and 6.4\%, 5.6\%, 6\% AUPRC enhancements than the best competing method.

\vspace{0.1cm}
\noindent\textbf{Generalization to Medical Image Anomalies.}
When applied to medical image AD datasets, \coolname consistently outperforms SotA models in all few-shot settings. It is evident that all competing methods perform poorly except WinCLIP. Impressively, using only two normal image prompts, \coolname can obtain over 97.3\% in AUPRC on BrainMRI, despite it does not have any training on medical data. On average, \coolname surpass the best competing model  by 3.9\%, 3.4\%, 3.9\% in AUROC and 0.6\%, 1\%, 1\% in AUPRC for $K = 2, 4, 8$ settings, respectively. 

\begin{table*}[t]
\centering
\resizebox{0.75\textwidth}{!}{
\begin{tabular}{|ccc|cccc|cc|cccc|}
\hline
                                &                                        &                                        & \multicolumn{4}{c|}{}                                                                                                                                         & \multicolumn{2}{c|}{}                                                         & \multicolumn{4}{c|}{\textbf{Semantic Anomalies}}                                                                                                                                   \\ \cline{10-13} 
                                &                                        &                                        & \multicolumn{4}{c|}{\multirow{-2}{*}{\textbf{Industrial Defects}}}                                                                                            & \multicolumn{2}{c|}{\multirow{-2}{*}{\textbf{Medical Anomalies}}}             & \multicolumn{2}{c|}{\textbf{One-vs-all}}                                                           & \multicolumn{2}{c|}{\textbf{Multi-class}}                                     \\ \cline{4-13} 
\multirow{-3}{*}{\textbf{$T$}} & \multirow{-3}{*}{\textbf{$P$}} & \multirow{-3}{*}{\textbf{$I$}} & \textbf{ELPV}                         & \textbf{SDD}                          & \textbf{AITEX}                        & \textbf{VisA}                  & \textbf{BrainMRI}                     & \textbf{HeadCT}                       & \textbf{MNIST}                 & \multicolumn{1}{c|}{\textbf{CIFAR-10}}              & \textbf{MNIST}                & \textbf{CIFAR-10}            \\ \hline
\checkmark                    & $\times$                            & $\times$                             & 0.733                                 & 0.946                                 & 0.733                                 & 0.787                                 & 0.926                                 & 0.900                                 & 0.678                                 & \multicolumn{1}{c|}{{\color[HTML]{0000FF} \textbf{0.924}}} & 0.620                                 & 0.900                                 \\
$\times$                      & \checkmark                           & $\times$                             & 0.794                                 & 0.946                                 & 0.730                                 & {\color[HTML]{0000FF} \textbf{0.843}}                                 & 0.859                                 & 0.829                                 & {\color[HTML]{0000FF} \textbf{0.890}} & \multicolumn{1}{c|}{0.850}                                 & 0.504                                 & 0.894                                 \\
$\times$                      & $\times$                             & \checkmark                           & 0.791                                 & 0.931                                 & {\color[HTML]{FF0000} \textbf{0.796}} & 0.808                                 & 0.898                                 & 0.906                                 & 0.694                                 & \multicolumn{1}{c|}{0.704}                                 & 0.612                                 & 0.712                                 \\
\checkmark                    & \checkmark                           & $\times$                            & 0.796                                 & {\color[HTML]{0000FF} \textbf{0.947}} & 0.711                                 & 0.840                                 & 0.938                                 & 0.919                                 & 0.887                                 & \multicolumn{1}{c|}{{\color[HTML]{0000FF} \textbf{0.924}}} & 0.622                                 & {\color[HTML]{0000FF} \textbf{0.920}} \\
\checkmark                    & $\times$                             & \checkmark                           & 0.783                                 & 0.938                                 & 0.756                                 & 0.792                                 & 0.932                                 & {\color[HTML]{0000FF} \textbf{0.924}} & 0.601                                 & \multicolumn{1}{c|}{0.800}                                 & {\color[HTML]{0000FF} \textbf{0.627}} & 0.905                                 \\
$\times$                      & \checkmark                           & \checkmark                           & {\color[HTML]{0000FF} \textbf{0.816}} & 0.945                                 & {\color[HTML]{0000FF} \textbf{0.785}} & {\color[HTML]{FF0000} \textbf{0.856}} & 0.952 & 0.918                                 & 0.760                                 & \multicolumn{1}{c|}{0.765}                                 & 0.618                                 & 0.821                                 \\
\checkmark                    & \checkmark                           & \checkmark                           & {\color[HTML]{FF0000} \textbf{0.839}} & {\color[HTML]{FF0000} \textbf{0.972}} & 0.761                                 & {\color[HTML]{FF0000} \textbf{0.856}} & {\color[HTML]{FF0000} \textbf{0.973}} & {\color[HTML]{FF0000} \textbf{0.929}} & {\color[HTML]{FF0000} \textbf{0.892}} & \multicolumn{1}{c|}{{\color[HTML]{FF0000} \textbf{0.935}}} & {\color[HTML]{FF0000} \textbf{0.635}} & {\color[HTML]{FF0000} \textbf{0.924}} \\ \hline
\multicolumn{3}{|c|}{\textbf{concatenation}}                                                                         & 0.809                                 & 0.951                                 & 0.708                                 & 0.832                                 & {\color[HTML]{0000FF} \textbf{0.955}}                                 & 0.913                                 & 0.819                                 & \multicolumn{1}{c|}{0.863}                                 & 0.597                                 & 0.823                                 \\
\multicolumn{3}{|c|}{\textbf{average}}                                                                            & 0.793                                 & 0.940                                 & 0.720                                 & 0.809                                 & 0.941                                 & 0.922                                 & 0.776                                 & \multicolumn{1}{c|}{0.881}                                 & 0.567                                 & 0.834                                 \\ \hline
\end{tabular}}
\caption{AUROC results for ablation study under two-shot setting. Best results and the second-best results are respectively in {\color[HTML]{FF0000} \textbf{red}} and {\color[HTML]{0000FF} \textbf{blue}}. The results for VisA, and the one-vs-all settings of MNIST and CIFAR-10 represent an average result across their respective data subsets}
\label{ab_study}
\vspace{-0.5cm}
\end{table*}

\begin{figure}[t]
    \centering
    \includegraphics[width=0.45\textwidth]{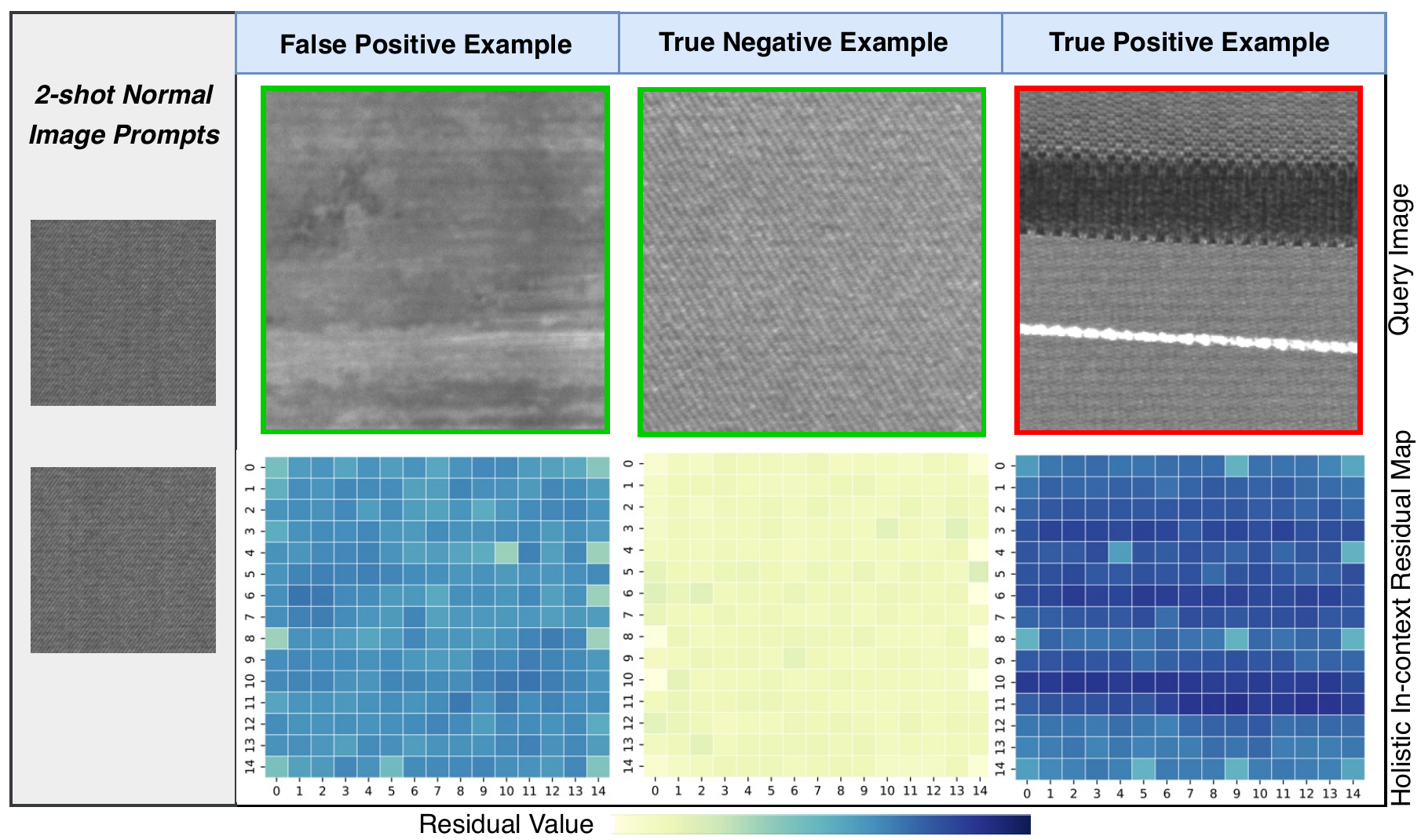}
    \caption{Visualization of query images $x_t$ and their holistic in-context residual maps $\mathbf{M}_{x_t}^+$. {\color{green}Green} and {\color{red}Red} frames indicate normal and abnormal images respectively. Deeper colors in the residual maps represent larger residual values.}
    \label{fig:qa}
    \vspace{-0.5cm}
\end{figure}

\vspace{0.1cm}
\noindent\textbf{Generalization to Semantic Anomalies under Both One-vs-all and Multi-class Settings.}
On detecting semantic anomalies, \coolname again consistently surpasses all SotA models. Remarkably, \coolname can obtain 90+\% in AUROC when the previous SotA methods can obtain 50\%-65\% AUROC only, showcasing highly promising GAD performance. Notably, WinCLIP achieves good performance on CIFAR-10. In contrast, CoOp experiences a notable decline, presumably losing crucial semantic knowledge when adapting to the auxiliary data that diverges significantly from the semantic AD task. 
Overall, our \coolname achieves the best performance with up to 8.2\%, 5.1\%, 4.4\% AUROC and 2.3\%, 1.9\%, 2.5\% AUPRC improvement compared to the best contender on $K = 2, 4, 8$ settings, respectively.

\subsection{Why Does \textbf{\coolname} Generalize Well?}
\vspace{0.1cm}
\noindent\textbf{Ablation Study.}
We examine the contribution of three key components of \coolname on the generalization:
text prompt-guided features ($T$), patch-level residuals ($P$), and image-level residuals ($I$), as well as their combinations. The results are reported in Table~\ref{ab_study}. The experiment results indicate that for industrial defect AD datasets, visual residual features play a more significant role compared to text prompt-based features, particularly on datasets like ELPV~\cite{deitsch2019automatic}, SDD~\cite{tabernik2020segmentation}, and AITEX~\cite{silvestre2019public}. On the medical image AD datasets, both visual residuals and textual knowledge contribute substantially to performance enhancement, exhibiting a complementary relation. On semantic AD datasets, the results are dominantly influenced by patch-level residuals and/or text prompt-based features. Importantly, our three components are generally mutually complementary, resulting in the superior detection generalization across the datasets.

\vspace{0.1cm}
\noindent\textbf{Significance of In-context Residual Learning.}
To assess the importance of learning the residuals in \coolname, we experiment with two alternative operations in both multi-layer patch-level and image-level residual learning: replacing the residual operation with 1) a \textbf{concatenation} operation and 2) an \textbf{average} operation, with all the other components of \coolname fixed. As shown in Table~\ref{ab_study}, the in-context residual learning generalizes much better than the other two alternative ways, significantly enhancing the model's performance in GAD across three distinct domains.

\vspace{-0.1cm}
\subsection{Failure Cases}
To understand the results of \coolname better, we provide visualization results illustrating both successful detection and failures by \coolname on ELPV \cite{deitsch2019automatic}. As depicted in Figure~\ref{fig:qa}, the incorrectly identified anomalies (False Positive example) shows substantially texture difference compared to the two normal image prompts, leading to similarly large residual values as that for a True Positive example. In contrast, when the query image resembles the normal image prompts well, the residual values are clearly very small, as shown by the True Negative example. These cases may be remedied when the image prompts include similar normal images as the falsely identified anomaly. This failure case exemplifies the challenge of GAD using only a few image prompts.

\setlength{\intextsep}{0pt}%
\setlength{\columnsep}{8pt}%
\begin{wrapfigure}[12]{l}{4.5cm}
\hspace{-0.65cm}
\includegraphics[height=0.18\textwidth]{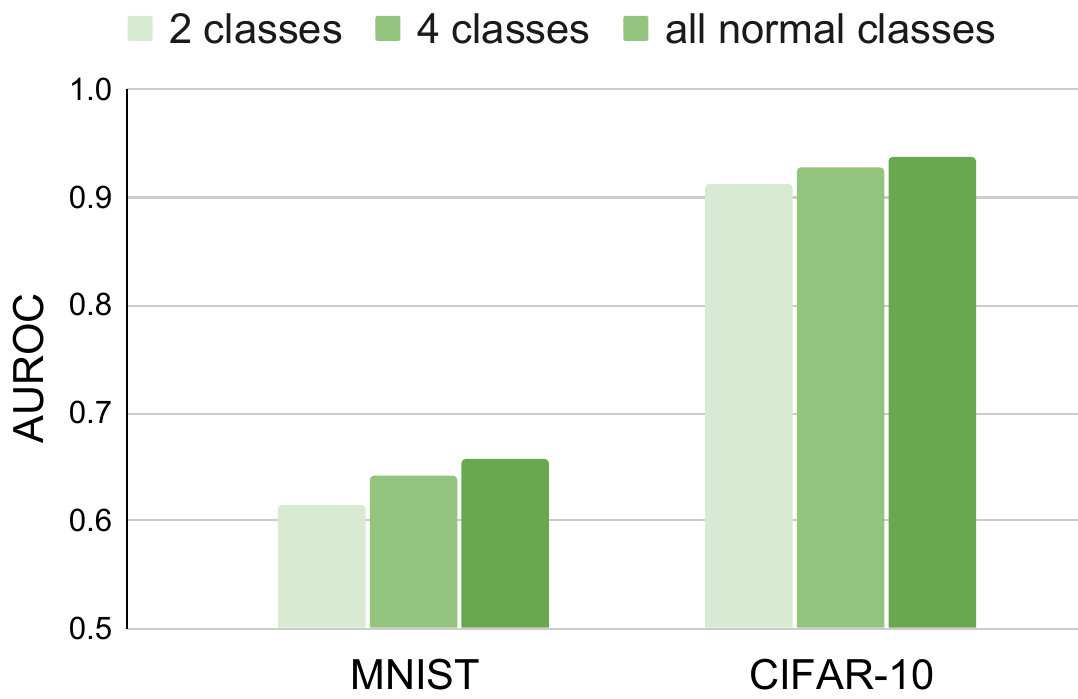}
\vspace{-0.3cm}
\caption{AUROC of \coolname with sample prompts from varying numbers of normal classes.}
\label{utility}
\end{wrapfigure}
\vspace{0.1cm}

This case is related to the diversity of few-shot normal sample prompts, which is particularly important when the normal data is complex, \eg, having multiple different normal patterns in our multi-class protocol. 

To rigorously investigate this problem, we evaluate the performance of the two datasets under the multi-class protocol, with varying numbers of normal classes included in the normal image prompts in the eight-shot setting. As shown in Figure~\ref{utility}, the performance of \coolname continually improves when the prompt sets have samples from more normal classes.

\section{Conclusion}
In this work we introduce a GAD task to evaluate the generalization capability of AD methods in identifying anomalies across various scenarios without any training on the target datasets. This is the first study dedicated to a generalist approach to anomaly detection, encompassing industrial defects, medical anomalies, and semantic anomalies. Then we propose an approach, called \coolname, to addressing this problem under a few-shot setting. \coolname achieves a superior GAD generalization by holistic in-context residual learning. Extensive experiments are performed on nine AD datasets to establish a GAD evaluation benchmark for the aforementioned three popular AD tasks, on which \coolname significantly and consistently outperforms SotA competing models across multiple few-shot settings.

{
    \small
    \bibliographystyle{ieeenat_fullname}
    \bibliography{main}
}
\appendix
\section{Dataset Details}
\subsection{Data Statistics of Training and Testing}
We conduct extensive experiments on nine real-world Anomaly Detection (AD) datasets, including five industrial defect inspection dataset (MVTec AD~\cite{bergmann2019mvtec}, VisA~\cite{zou2022spot}, ELPV~\cite{deitsch2019automatic}, SDD~\cite{tabernik2020segmentation}, AITEX~\cite{silvestre2019public}), two medical image datasets (BrainMRI~\cite{salehi2021multiresolution}, HeadCT~\cite{salehi2021multiresolution}), and two semantic anomaly detection datasets: MNIST~\cite{lecun1998gradient} and CIFAR-10~\cite{krizhevsky2009cifar} under both one-vs-all and multi-class protocols~\cite{Cao_2023_ICCV}. 

To assess the Genealist Anomaly Dtection (GAD) performance, the full dataset of MVTec AD, including both training set and test set, is used as the auxiliary training data, on which AD models are trained, and they are subsequently evaluated on the test set of the other eight datasets without any further training. We train the model on the full dataset of VisA when evaluating the performance on MVTec AD. The few-shot normal prompts for the target data are randomly sampled from the training set of target datasets and remain the same for all models for fair comparison. Table~\ref{train} provides the data statistics of MVTec AD and VisA, while Table~\ref{test} shows the test set statistics of the rest datasets.

\begin{table}[!t]
\centering
\resizebox{0.45\textwidth}{!}{
\begin{tabular}{|c|c|c|c|cc|}
\hline
\multirow{2}{*}{\textbf{Dataset}}   & \multirow{2}{*}{\textbf{Subset}} & \multirow{2}{*}{\textbf{Type}} & \textbf{Original Training} & \multicolumn{2}{c|}{\textbf{Original Test}} \\ \cline{4-6} 
                                    &                                  &                                & \textbf{Normal}            & \textbf{Normal}      & \textbf{Anomaly}     \\ \hline
\multirow{15}{*}{\textbf{MVTec AD}} & \textbf{Carpet}                  & Texture                        & 280                        & 28                   & 89                   \\
                                    & \textbf{Grid}                    & Texture                        & 264                        & 21                   & 57                   \\
                                    & \textbf{Leather}                 & Texture                        & 245                        & 32                   & 92                   \\
                                    & \textbf{Tile}                    & Texture                        & 230                        & 33                   & 83                   \\
                                    & \textbf{Wood}                    & Texture                        & 247                        & 19                   & 60                   \\
                                    & \textbf{Bottle}                  & Object                         & 209                        & 20                   & 63                   \\
                                    & \textbf{Capsule}                 & Object                         & 219                        & 23                   & 109                  \\
                                    & \textbf{Pill}                    & Object                         & 267                        & 26                   & 141                  \\
                                    & \textbf{Transistor}              & Object                         & 213                        & 60                   & 40                   \\
                                    & \textbf{Zipper}                  & Object                         & 240                        & 32                   & 119                  \\
                                    & \textbf{Cable}                   & Object                         & 224                        & 58                   & 92                   \\
                                    & \textbf{Hazelnut}                & Object                         & 391                        & 40                   & 70                   \\
                                    & \textbf{Metal\_nut}              & Object                         & 220                        & 22                   & 93                   \\
                                    & \textbf{Screw}                   & Object                         & 320                        & 41                   & 119                  \\
                                    & \textbf{Toothbrush}              & Object                         & 60                         & 12                   & 30                   \\ \hline
\multirow{12}{*}{\textbf{VisA}}     & \textbf{candle}                  & Object                         & 900                        & 100                  & 100                  \\
                                    & \textbf{capsules}                & Object                         & 542                        & 60                   & 100                  \\
                                    & \textbf{cashew}                  & Object                         & 450                        & 50                   & 100                  \\
                                    & \textbf{chewinggum}              & Object                         & 453                        & 50                   & 100                  \\
                                    & \textbf{fryum}                   & Object                         & 450                        & 50                   & 100                  \\
                                    & \textbf{macaroni1}               & Object                         & 900                        & 100                  & 100                  \\
                                    & \textbf{macaroni2}               & Object                         & 900                        & 100                  & 100                  \\
                                    & \textbf{pcb1}                    & Object                         & 904                        & 100                  & 100                  \\
                                    & \textbf{pcb2}                    & Object                         & 901                        & 100                  & 100                  \\
                                    & \textbf{pcb3}                    & Object                         & 905                        & 101                  & 100                  \\
                                    & \textbf{pcb4}                    & Object                         & 904                        & 101                  & 100                  \\
                                    & \textbf{pipe\_fryum}             & Object                         & 450                        & 50                   & 100                  \\ \hline
\end{tabular}}
\caption{Data statistics of MVTec AD and VisA. When training GAD models with MVTec AD or VisA datasets, we utilize their complete datasets, including both training and test data. In contrast, for testing GAD models, only the test sets of MVTec AD or VisA are employed for inference.}
\label{train}
\end{table}

\begin{table}[!t]
\centering
\resizebox{0.4\textwidth}{!}{
\begin{tabular}{|c|c|c|cc|}
\hline
\multirow{2}{*}{\textbf{Dataset}}   & \multirow{2}{*}{\textbf{Subset}} & \multirow{2}{*}{\textbf{Type}} & \multicolumn{2}{c|}{\textbf{Test set}} \\
                                    &                                  &                                & \textbf{Normal}   & \textbf{Anomaly}   \\ \hline
\multirow{11}{*}{\textbf{MNIST}}    & \textbf{0}                       & Semantical                     & 980               & 9020               \\
                                    & \textbf{1}                       & Semantical                     & 1135              & 8865               \\
                                    & \textbf{2}                       & Semantical                     & 1,032             & 8,968              \\
                                    & \textbf{3}                       & Semantical                     & 1,010             & 8,990              \\
                                    & \textbf{4}                       & Semantical                     & 982               & 9019               \\
                                    & \textbf{5}                       & Semantical                     & 892               & 9108               \\
                                    & \textbf{6}                       & Semantical                     & 958               & 9042               \\
                                    & \textbf{7}                       & Semantical                     & 1028              & 8972               \\
                                    & \textbf{8}                       & Semantical                     & 974               & 9026               \\
                                    & \textbf{9}                       & Semantical                     & 1009              & 8991               \\
                                    & \textbf{even\_number}            & Semantical                     & 4926              & 5074               \\ \hline
\multirow{11}{*}{\textbf{CIFAR-10}} & \textbf{airplane}                & Semantical                     & 1000              & 9000               \\
                                    & \textbf{automobile}              & Semantical                     & 1000              & 9000               \\
                                    & \textbf{bird}                    & Semantical                     & 1000              & 9000               \\
                                    & \textbf{cat}                     & Semantical                     & 1000              & 9000               \\
                                    & \textbf{deer}                    & Semantical                     & 1000              & 9000               \\
                                    & \textbf{dog}                     & Semantical                     & 1000              & 9000               \\
                                    & \textbf{frog}                    & Semantical                     & 1000              & 9000               \\
                                    & \textbf{horse}                   & Semantical                     & 1000              & 9000               \\
                                    & \textbf{ship}                    & Semantical                     & 1000              & 9000               \\
                                    & \textbf{truck}                   & Semantical                     & 1000              & 9000               \\
                                    & \textbf{animal}                  & Semantical                     & 6000              & 4000               \\ \hline
\textbf{ELPV}                       & \textbf{-}                       & Texture                        & 377               & 715                \\ \hline
\textbf{SDD}                        & \textbf{-}                       & Texture                        & 286               & 54                 \\ \hline
\textbf{AITEX}                      & \textbf{-}                       & Texture                        & 564               & 183                \\ \hline
\textbf{BrainMRI}                   & \textbf{-}                       & Medical                        & 25                & 155                \\ \hline
\textbf{HeadCT}                     & \textbf{-}                       & Medical                        & 25                & 100                \\ \hline
\end{tabular}}
\caption{Data statistics of seven AD datasets for inference. These datasets are exclusively used for inference purposes, hence only the details of the test sets are provided.}
\label{test}
\end{table}

\subsection{Industrial Defect Inspection Datasets}
\vspace{0.1cm}
\noindent\textbf{MVTec AD}~\cite{bergmann2019mvtec} is a widely-used dataset that enables researchers to benchmark the performance of anomaly detection methods in the context of industrial inspection applications. The dataset includes over 5,000 images that are divided into 15 object and texture categories. Each category contains a training set of anomaly-free images, as well as a test set that includes images with both defects and defect-free images. 

\vspace{0.1cm}
\noindent\textbf{VisA}~\cite{zou2022spot} consists of 10,821 high-resolution color images (9,621 normal and 1,200 anomalous samples) covering 12 objects in 3 domains, making it the largest industrial anomaly detection dataset to date. Both image and pixel-level labels are provided. The anomalous images contain various flaws, including surface defects such as scratches, dents, color spots or crack, and structural defects like misplacement or missing parts.

\vspace{0.1cm}
\noindent\textbf{ELPV}~\cite{deitsch2019automatic} is a collection of 2,624 high-resolution grayscale images of solar cells extracted from photovoltaic modules. These images were extracted from 44 different solar modules, and include both intrinsic and extrinsic defects known to reduce the power efficiency of solar modules. In our study, we only use its test set for evaluation.

\vspace{0.1cm}
\noindent\textbf{SDD}~\cite{tabernik2020segmentation} is a collection of images captured in a controlled industrial environment, using defective production items as the subject. The dataset includes 52 images with visible defects and 347 product images without any defects. In our study, we only use its test set for evaluation.

\vspace{0.1cm}
\noindent\textbf{AITEX}~\cite{silvestre2019public} is a textile fabric database that comprises 245 images of 7 different fabrics, including 140 defect-free images (20 for each type of fabric) and 105 images with various types of defects. We only use its test set for evaluation.

\subsection{Medical Anomaly Detection Datasets}

\vspace{0.1cm}
\noindent\textbf{BrainMRI}~\cite{salehi2021multiresolution} is a dataset for brain tumor detection obtained from magnetic resonance imaging (MRI) of the brain. In our study, we only use its test set for evaluation.

\vspace{0.1cm}
\noindent\textbf{HeadCT}~\cite{salehi2021multiresolution} is a dataset consisting of 100 normal head CT slices and 100 slices with brain hemorrhage, without distinction between the types of hemorrhage. Each slice is from a different person, providing a diverse set of images for researchers to develop and test algorithms for hemorrhage detection and classification in medical imaging applications. In our study, we only use its test set for evaluation.

\subsection{Semantic Anomaly Detection Datasets}

\vspace{0.1cm}
\noindent\textbf{MNIST}~\cite{lecun1998gradient} encompasses 70,000 grayscale images of handwritten digits. It serves as a semantic AD dataset in our work, where we utilize its original test set to construct test sets of one-vs-all and multi-class settings. Under the one-vs-all protocol, one of the ten classes is used as normal, with the other classes treated as abnormal; while under the multi-class protocol, images of even-number classes are treated as normal, with the images of the other classes are considered as anomalies. In this case, the category-level label is set to `even\_number'. 

\vspace{0.1cm}
\noindent\textbf{CIFAR-10}~\cite{krizhevsky2009cifar} consists of 60,000 colour images in 10 classes, with 6,000 images per class. There are 50,000 training images and 10,000 test images. It serves as a semantic AD dataset in our work, where we utilize its original test set to construct test sets of one-vs-all and multi-class settings. 
Under the one-vs-all protocol, one of the 10 classes is used as normal, with the other classes treated as abnormal; while under the multi-class protocol, images of animal-related classes are treated as normal, with the images of the other classes are considered as anomalies. In this case, the category-level label is set to `animal'.

\section{Implementation Details}
\subsection{Data Pre-processing}
By default, for all CLIP-based models, including WinCLIP~\cite{jeong2023winclip}, CoOp~\cite{zhou2022conditional}, and \coolname, we adopt the same CLIP implementation, OpenCLIP~\cite{ilharco2021openclip}, and its public pre-trained backbone ViT-B/16+ in our experiments. Our data preprocessing aligns with OpenCLIP across all datasets. Specifically, this involves channel-wise standardization using a predefined mean 
and standard deviation 
after scaling RGB images to the range of [0, 1], followed by bicubic resizing based on Pillow library. In addition, we resize the input resolution to 240$\times$240 to match ViT-B/16+. This resizing is also applied to other baseline models for fair comparison, while retaining their original data preprocessing methods (if there are any).

\subsection{Network Architectures}
In our experiments, the parameters of visual encoder and text encoder of ViT-B/16+ are kept frozen. This model, while being similar in depth to ViT-B/16, increases the dimensions of image embeddings(768 $\rightarrow$ 896), text embedding(512 $\rightarrow$ 640) and input resolution (224$\times$224 $\rightarrow$ 240$\times$240). For the learnable components, to align with ViT-B/16+'s dimensions, the adapter $\psi$ has input and output dimensions set to 896, including a 224-unit hidden layer with ReLU activation. The image-level anomaly classification learner $\eta$ takes in-context image-level residual features $F_x$as input and yields a one-dimensional prediction, where $\eta$ has two hidden layers with 128 and 64 units respectively. The holistic anomaly scoring model $\phi$ incorporates two hidden layers, projecting a 225-dimensional in-context residual map to generate a final single-dimensional anomaly score.

\begin{table}[!t]
\centering
\resizebox{0.45\textwidth}{!}{
\begin{tabular}{|c|c|}
\hline
\multirow{5}{*}{\textbf{Normal Examples}}    & `\textit{a photo of a flawless [c] for visual inspection.}'  \\
                                             & `\textit{a cropped photo of a perfect [c].}'                 \\
                                             & `\textit{a blurry photo of the [c]  without defect.}'        \\
                                             & `\textit{a dark photo of the unblemished [c].}'              \\
                                             & `\textit{a jpeg corrupted photo of a [c] without flaw.}'     \\ \hline
\multirow{5}{*}{\textbf{Abnormal Examples}} & `\textit{a photo of a [c] with flaw for visual inspection.}' \\
                                             & `\textit{a cropped photo of a [c] with damage.}'             \\
                                             & `\textit{a blurry photo of the [c]  with defect.}'           \\
                                             & `\textit{a dark photo of the  [c] with flaw.}'               \\
                                             & `\textit{a jpeg corrupted photo of a [c] with defect.}'      \\ \hline
\end{tabular}}
\caption{Examples of normal and abnormal text prompts used in \coolname and WinCLIP. \textit{[c]} represents a class label.}
\label{example}
\end{table}

\subsection{Details of Text prompts}
The text prompts used in our work are based on the same main text and ensemble strategy to WinCLIP~\cite{jeong2023winclip} except CIFAR-10~\cite{zou2022spot}. Table~\ref{example} provides several normal and abnormal examples of text prompts used in \coolname (see \cite{jeong2023winclip} for the full list of the text prompts). The WinCLIP prompts fail to work for natural images like CIFAR-10, so the normal and abnormal text prompts of CIFAR-10 are designed as `\textit{a photo of [c] for anomaly detection.}' and `\textit{a photo without [c] for anomaly detection.}', respectively, which are used in both WinCLIP and \coolname on CIFAR-10. Here, \textit{[c]} represents a category-level label, \eg, airplane.

\subsection{Implementation of Comparison Methods}
For the results of competing methods, we re-implemented SPADE, PaDiM, and WinCLIP, while using the official implementations of PatchCore, RegAD, and CoOp. Differing from SPADE's original $K=50$ setup, we use $K=2, 4, 8$ nearest neighbors to match the few-shot setting. For PaDiM, we select the wide\_resnet50\_2 model, pretrained on ImageNet, as the feature extractor. To ensure fair empirical comparison, we apply the same image prompts as used in \coolname across all methods. All reported results are the average of three independent runs, each with a different random seed.

\section{Detailed Empirical Results}

\subsection{Complexity of \coolname vs. Other CLIP-based Methods}
The number of parameters and per-image inference time for CLIP-based methods are shown in Table~\ref{comp}. 

Our \coolname and CoOp involve additional training on auxiliary data compared to the training-free method, \ie, WinCLIP. This results in extra trainable parameters during the training phase.
However, the extra time consumption leads to significant performance enhancements in \coolname, and furthermore, the training can be taken offline, so its 
computation overhead is generally negligible. 

Additionally, as Table~\ref{comp} shows, \coolname achieves faster inference compared to WinCLIP's multi-scale few-shot anomaly scoring approach. Although CoOp adopts a similar few-shot anomaly scoring method as WinCLIP, it gains better efficiency by avoiding the ensemble text prompt strategy in WinCLIP. Result indicates the effectiveness of the \coolname framework in enhancing the base model's generalization ability.

\begin{table}[!t]
\centering
\resizebox{0.45\textwidth}{!}{
\begin{tabular}{|c|c|c|}
\hline
\textbf{Method}  & \textbf{Number of Parameters} & \textbf{Inference Time (ms)} \\ \hline
\textbf{CoOp}    & 6,400                & 197.3±5.6                    \\ 
\textbf{WinCLIP} & 0                   & 227.5±0.7                    \\ 
\textbf{Ours (\coolname)}    & 334,916              & 81.7±1.4                     \\ \hline
\end{tabular}}
\caption{Number of Parameters and Per-image Inference time.}
\label{comp}
\end{table}

\subsection{Comparison with Traditional Medical Anomaly Detection Methods}
Even though our comparison is focused on detectors of similar generalist detection capabilities, we compare five recent AD methods specifically designed for medical images on two other medical datasets (Brain Tumor MRI\footnote{The dataset is available at \url{https://www.kaggle.com/datasets/masoudnickparvar/brain-tumor-mri-dataset}.} and LAG~\cite{li2019attention}) in Table~\ref{mad}. The results of traditional medical anomaly detection methods are from Cai \etal~\cite{Cai_2023} based on \textbf{full-shot} one-class classification setting. It should be noted that better performance is gained if \textbf{extra anomaly image data} is used, but it would be very unfair comparison to our method that uses only \textbf{8-shot} normal images. 

As shown in Table~\ref{mad}, our method outperforms all competing models on almost all cases, despite the fact that our method uses only 8-shot normal images for prompting and does not require any training on the medical data whereas the medical image AD methods require extensive training on a large set of normal medical images, indicating the superior generalized AD capability of our model.

\begin{table}[!t]
\centering
\resizebox{0.45\textwidth}{!}{
\begin{tabular}{|c|c|cc|cc|}
\hline
\multicolumn{1}{|l|}{\multirow{2}{*}{}}                  & \multirow{2}{*}{\textbf{Dataset}} & \multicolumn{2}{c|}{\textbf{Brain Tumor MRI}} & \multicolumn{2}{c|}{\textbf{LAG}} \\ \cline{3-6} 
\multicolumn{1}{|l|}{}                                   &                                   & AUR                   & PR                    & AUR             & PR              \\ \hline
\multirow{5}{*}{\textbf{Traditional Medical AD}} & \textbf{FPI}                      & 0.831                 & 0.789                 & 0.543           & 0.556           \\
                                                         & \textbf{PII}                      & 0.843                 & 0.805                 & 0.610           & 0.607           \\
                                                         & \textbf{F-AnoGAN}                 & 0.825                 & 0.743                 & 0.842           & 0.775           \\
                                                         & \textbf{AEU}                      & 0.940                 & 0.890                 & 0.813           & 0.789           \\
                                                         & \textbf{AEU+DDAD}                 & 0.942                 & 0.919                 & \textbf{0.860}  & 0.840           \\ \hline
\multirow{2}{*}{\textbf{Generalist AD}}          & \textbf{WinCLIP}                  & 0.779                 & 0.878                 & 0.571           & 0.731           \\
                                                         & \textbf{Ours (\coolname)}                     & \textbf{0.951}        & \textbf{0.968}        & 0.832           & \textbf{0.880}  \\ \hline
\end{tabular}}
\caption{Comparison with Traditional Medical AD Methods.}
\label{mad}
\end{table}

\subsection{Full Results on VisA and MVTec AD}
Table~\ref{visa} presents detailed comparison results of \coolname against six SotA methods across each category of the VisA dataset. Overall, \coolname markedly surpasses all competitors in every case within the three few-shot settings. We observe a general improvement in performance across all methods with an increase in the number of few-shot image prompts.

Similarly, Table~\ref{mvtecad} details the results of \coolname and six SotA methods across each category of the MVTec AD dataset. \coolname again consistently outperforms all baseline models in all few-shot settings.

\begin{table*}[!ht]
\centering
\resizebox{\textwidth}{!}{
\begin{tabular}{|cc|ccccccc|ccccccc|}
\hline
\multicolumn{2}{|c}{}                                                           & \multicolumn{7}{|c|}{\textbf{AUROC}}                                                                                                                                                                          & \multicolumn{7}{c|}{\textbf{AUPRC}}                                                                                                                                                                          \\ \cline{3-16} 
\multicolumn{2}{|c|}{\multirow{-2}{*}{\textbf{Data subset}}}                     & \textbf{SPADE}       & \textbf{PaDiM}       & \textbf{Patchcore}   & \textbf{RegAD}       & \textbf{CoOp}        & \textbf{WinCLIP}                            & \textbf{Ours (\coolname)}                               & \textbf{SPADE}       & \textbf{PaDiM}       & \textbf{Patchcore}   & \textbf{RegAD}       & \textbf{CoOp}        & \textbf{WinCLIP}                            & \textbf{Ours (\coolname)}                               \\ \hline
\multicolumn{1}{|c|}{}                                   & \textbf{candle}      & 0.840{\scriptsize±0.057}          & 0.876{\scriptsize±0.017}          & 0.935{\scriptsize±0.013}          & 0.455{\scriptsize±0.042}          & 0.943{\scriptsize±0.019}          & 0.974{\scriptsize±0.007}                                 & 0.916{\scriptsize±0.006}                                 & 0.889{\scriptsize±0.033}          & 0.839{\scriptsize±0.015}          & 0.940{\scriptsize±0.021}          & 0.461{\scriptsize±0.038}          & 0.950{\scriptsize±0.015}          & 0.974{\scriptsize±0.006}                                 & 0.920{\scriptsize±0.008}                                 \\
\hline
\end{tabular}}
\caption{Fine-grained AUROC and AUPRC results(mean±std) on VisA datasets under various few-shot AD settings. Best results and the second-best results are respectively highlighted in {\color[HTML]{FF0000} \textbf{red}} and {\color[HTML]{0000FF} \textbf{blue}}.}
\label{visa}
\end{table*}

\begin{table*}[!ht]
\centering
\resizebox{\textwidth}{!}{
\begin{tabular}{|cc|ccccccc|ccccccc|}
\hline
\multicolumn{2}{|c|}{}                                                         & \multicolumn{7}{c|}{\textbf{AUROC}}                                                                                                                                                                          & \multicolumn{7}{c|}{\textbf{AUPRC}}                                                                                                                                                                          \\ \cline{3-16} 
\multicolumn{2}{|c|}{\multirow{-2}{*}{\textbf{Data subset}}}                   & \textbf{SPADE}       & \textbf{PaDiM}       & \textbf{Patchcore}   & \textbf{RegAD}       & \textbf{CoOp}        & \textbf{WinCLIP}                            & \textbf{Ours (\coolname)}                               & \textbf{SPADE}       & \textbf{PaDiM}       & \textbf{Patchcore}   & \textbf{RegAD}       & \textbf{CoOp}        & \textbf{WinCLIP}                            & \textbf{Ours (\coolname)}                               \\ \hline
\multicolumn{1}{|c|}{}                                   & \textbf{carpet}     & 0.976{\scriptsize±0.008 }         & 0.992{\scriptsize±0.005}          & 0.996{\scriptsize±0.003}          & 0.601{\scriptsize±0.051}          & 1.000{\scriptsize±0.000}          & 0.998{\scriptsize±0.002 }                                & 0.997{\scriptsize±0.003 }                                & 0.962{\scriptsize±0.005}          & 0.998{\scriptsize±0.003}          & 0.999{\scriptsize±0.001}          & 0.853{\scriptsize±0.023}          & 1.000{\scriptsize±0.000}          & 1.000{\scriptsize±0.000}                                 & 0.999{\scriptsize±0.001 }                                \\
\hline
\end{tabular}}
\caption{Fine-grained AUROC and AUPRC results(mean±std) on MVTec AD datasets under various few-shot AD settings. Best results and the second-best results are respectively highlighted in {\color[HTML]{FF0000} \textbf{red}} and {\color[HTML]{0000FF} \textbf{blue}}.}
\label{mvtecad}
\end{table*}

\begin{table*}[!ht]
\centering
\resizebox{\textwidth}{!}{
\begin{tabular}{|cc|cccccc|cccccc|}
\hline
\multicolumn{2}{|c|}{}                                                   & \multicolumn{6}{c|}{\textbf{AUROC}}                                                                                                                                                                          & \multicolumn{6}{c|}{\textbf{AUPRC}}                                                                                                                                                                          \\ \cline{3-14} 
\multicolumn{2}{|c|}{\multirow{-2}{*}{\textbf{Data subset}}}             & \textbf{SPADE}       & \textbf{Patchcore}                          & \textbf{RegAD}       & \textbf{CoOp}        & \textbf{WinCLIP}                            & \textbf{Ours (\coolname)}                               & \textbf{SPADE}       & \textbf{Patchcore}                          & \textbf{RegAD}       & \textbf{CoOp}        & \textbf{WinCLIP}                            & \textbf{Ours (\coolname)}                               \\ \hline
\multicolumn{1}{|c|}{}                                   & \textbf{0}    & 0.772{\scriptsize±0.174}          & 0.803{\scriptsize±0.028 }                                & 0.615{\scriptsize±0.037}          & 0.508{\scriptsize±0.031}          & 0.801{\scriptsize±0.026}                                 & 0.970{\scriptsize±0.004}                                 & 0.959{\scriptsize±0.036}          & 0.950{\scriptsize±0.003}                                 & 0.923{\scriptsize±0.013}          & 0.927{\scriptsize±0.004}          & 0.954{\scriptsize±0.004}                                 & 0.995{\scriptsize±0.000 }                                \\
\hline
\end{tabular}}
\caption{Fine-grained AUROC and AUPRC results(mean±std) on One-vs-all protocol of MNIST datasets under various few-shot AD settings. Best results and the second-best results are respectively highlighted in {\color[HTML]{FF0000} \textbf{red}} and {\color[HTML]{0000FF} \textbf{blue}}.}
\label{mnist}
\end{table*}

\begin{table*}[!ht]
\centering
\resizebox{\textwidth}{!}{
\begin{tabular}{|cc|cccccc|cccccc|}
\hline
\multicolumn{2}{|c|}{}                                                         & \multicolumn{6}{c|}{\textbf{AUROC}}                                                                                                                                                   & \multicolumn{6}{c|}{\textbf{AUPRC}}                                                                                                                                                   \\ \cline{3-14} 
\multicolumn{2}{|c|}{\multirow{-2}{*}{\textbf{Data subset}}}                   & \textbf{SPADE}       & \textbf{Patchcore}   & \textbf{RegAD}       & \textbf{CoOp}        & \textbf{WinCLIP}                            & \textbf{Ours (\coolname)}                               & \textbf{SPADE}       & \textbf{Patchcore}   & \textbf{RegAD}       & \textbf{CoOp}        & \textbf{WinCLIP}                            & \textbf{Ours (\coolname)}                               \\ \hline
\multicolumn{1}{|c|}{}                                   & \textbf{airplane}   & 0.822{\scriptsize±0.053}          & 0.566{\scriptsize±0.065}          & 0.439{\scriptsize±0.043}          & 0.437{\scriptsize±0.014 }         & 0.917{\scriptsize±0.005}                                 & 0.938{\scriptsize±0.006 }                                & 0.974{\scriptsize±0.009}          & 0.915{\scriptsize±0.014}          & 0.865{\scriptsize±0.011}          & 0.886{\scriptsize±0.005  }        & 0.988{\scriptsize±0.002}                                 & 0.993{\scriptsize±0.001 }                                \\
\hline
\end{tabular}}
\caption{Fine-grained AUROC and AUPRC results(mean±std) on One-vs-all protocol of CIFAR-10 datasets under various few-shot AD settings. Best results and the second-best results are respectively highlighted in {\color[HTML]{FF0000} \textbf{red}} and {\color[HTML]{0000FF} \textbf{blue}}.}
\label{cifar10}
\end{table*}
 
\subsection{Full Results on One-vs-all Setting of Sementic Anomaly Detection Datasets}
Table~\ref{mnist} provides the detailed empirical comparison results of \coolname against five SotA methods using the one-vs-all protocol on the MNIST dataset. Similarly, Table~\ref{cifar10} details the performance of \coolname relative to five SotA methods under the one-vs-all protocol on the CIFAR-10 dataset. \coolname consistently surpasses all baseline methods in all few-shot settings on both datasets.

\end{document}